%% file: main.tex
\documentclass[11pt]{article}

\usepackage[preprint]{acl}

\usepackage{times}
\usepackage{latexsym}
\usepackage{natbib}

\usepackage[T1]{fontenc}

\usepackage[utf8]{inputenc}

\usepackage{microtype}

\usepackage{inconsolata}

\usepackage{graphicx}
\usepackage{subcaption}
\usepackage{tcolorbox}
\usepackage{multirow}
\usepackage{amsmath}
\usepackage{amssymb}
\usepackage{url}
\usepackage{xurl}
\usepackage{multirow}
\usepackage{booktabs}

\usepackage{amsmath}
\usepackage{amssymb}
\usepackage[dvipsnames,svgnames]{xcolor} 
\usepackage{tcolorbox}
\tcbuselibrary{skins}
\usepackage{algorithm}
\usepackage{algorithmic}
\usepackage{caption}

\definecolor{sedateBlue}{HTML}{16425B}
\definecolor{sedateTeal}{HTML}{0F8B8D}
\definecolor{sedateAmber}{HTML}{E07A1F}
\definecolor{sedateRed}{HTML}{C0392B}
\definecolor{sedateGreen}{HTML}{2E7D32}
\definecolor{sedateGray}{HTML}{6B7280}
\definecolor{sedatePaper}{HTML}{FBFAF7}
\definecolor{sedateEditBG}{HTML}{FFF4E6}
\definecolor{sedateFallBG}{HTML}{FDECEA}
 
\makeatletter
\renewcommand{\ALC@linenosize}{\footnotesize\color{sedateGray}}
\makeatother
 
\newcommand{\stagebadge}[2]{%
  \tcbox[on line, boxsep=1pt, left=3pt, right=3pt, top=1pt, bottom=1pt,
         colback=#1, colframe=#1, arc=2pt,
         fontupper=\scriptsize\bfseries\color{white}]{#2}}
 
\newtcolorbox[use counter=algorithm]{sedatebox}[1][]{%
  enhanced,
  float=t,
  colback=sedatePaper,
  colframe=sedateBlue,
  boxrule=0.9pt, arc=3.5pt,
  drop fuzzy shadow={opacity=0.35},
  left=6pt, right=6pt, top=6pt, bottom=5pt,
  title={Algorithm~\thetcbcounter\;#1},,
  fonttitle=\bfseries\color{white},
  coltitle=white,
  colbacktitle=sedateBlue,
  attach boxed title to top left={yshift=-2mm, xshift=4mm},
  boxed title style={arc=2.5pt, boxrule=0pt,
    interior style={left color=sedateBlue, right color=sedateTeal}},
}
\usepackage{newfloat}
\usepackage{listings}
\DeclareCaptionStyle{ruled}{labelfont=normalfont,labelsep=colon,strut=off} 
\floatstyle{ruled}
\newfloat{listing}{tb}{lst}{}
\floatname{listing}{Listing}

\usepackage{booktabs}
\usepackage{tcolorbox}
\tcbuselibrary{breakable,skins}
\usepackage{xcolor}
\definecolor{boxblue}{RGB}{31,78,121}
\definecolor{boxbackground}{RGB}{245,248,252}

\newcommand{\ours}[1]{\textsc{SeDATE}}

\newcommand{\am}[1]{\textcolor{red}{[AM: #1]}}
\usepackage{enumitem}
\usepackage{amssymb}
\usepackage{xcolor}

\definecolor{accent}{HTML}{8C1515}   
\definecolor{boxbg}{HTML}{FAF6F4}

\newtcolorbox{contribbox}[1][Contributions]{
  enhanced, breakable,
  colback=boxbg, colframe=accent,
  boxrule=0pt, leftrule=2.5pt,
  arc=0pt, outer arc=0pt,
  left=5pt, right=10pt, top=8pt, bottom=8pt,
  before upper={%
    {\color{accent}\small\bfseries\scshape #1}\par\nointerlineskip
    \vspace{1pt}{\color{accent!40}\rule{\linewidth}{0.4pt}}\par\vspace{4pt}%
  }
}
\tcbset{
  premiumbox/.style={
    enhanced, breakable,
    colback=boxbg, colframe=accent,
    boxrule=0pt, leftrule=2.5pt,
    arc=0pt, outer arc=0pt,
    left=5pt, right=6pt, top=6pt, bottom=6pt,
    before upper={%
      {\color{accent}\small\bfseries\scshape #1}\par\nointerlineskip
      \vspace{1pt}{\color{accent!40}\rule{\linewidth}{0.4pt}}\par\vspace{4pt}%
    }
  }
}

\newtcolorbox{researchgap}[1][Research Question]{premiumbox=#1}

\title{\ours{}: Do Large Language Models Share\\ an Aggression Circuit Across Languages?}

\input{authors}

\begin{document}
\maketitle

\begin{abstract}
As the applications of AI continue to expand across various multimodal and multilingual domains, ensuring safe and responsible deployment of these systems has become an important concern. A wide-range of interventions have been proposed to make AI responses safer and more reliable. They often incur substantial computational or annotation costs, or cannot be extended to multilingual settings. In this work, we propose \underline{Se}lective \underline{D}etection \underline{A}nd \underline{T}argeted \underline{E}diting (\ours{}), a training-free approach based on \textbf{\texttt{detect}-\texttt{edit}-\texttt{verify}} strategy. \ours{} first detects the responsible neurons, edits them as and when necessary and finally verifies the model responses; all without any additional model training. We evaluate \ours{} across 7 languages and 4 model families. We further evaluate its effectiveness across 4 out-of-distribution languages and 4 additional datasets. We show that \ours{} achieves substantially superior safety scores across all languages and models over the best-performing baseline while retaining the general capabilities and incurring negligible overhead. Our results are robust, with the trends remaining similar for out-of-distribution languages and datasets.\\
\textcolor{red}{\textbf{\textit{Warning: This paper contains text that some readers may find harmful or offensive.}}}
\end{abstract}


\section{Introduction}

\noindent Existing safety-alignment methods mainly rely on system prompts~\cite{sharma2026sysformer, xie2023defending}, supervised fine-tuning, preference optimization, or reinforcement learning from human feedback~\cite{wei2026jailbreak, banerjee2025navigating}. Although effective, training-based methods require substantial amounts of curated safety data and computational resources, while prompt-based defenses can remain vulnerable to adversarial reformulations~\cite{kumar2024certifying}. Inference-time interventions provide a lightweight alternative by directly modifying internal model representations without retraining the underlying parameters~\cite{hazra-etal-2024-safety}. However, most existing methods focus on steering model behaviour and provide limited understanding of whether harmful generations across different languages arise from shared or language-specific internal mechanisms~\cite{tang-etal-2024-language}. This motivates a fundamental question as follows.
\begin{researchgap}
\textit{Does aggressive generation rely on different neurons in different
languages, or is there a shared set of neurons that consistently
responds to aggressive content across languages?}
\end{researchgap}

\noindent Answering this question is important for both interpretability and safety control. If aggressive behaviour is partly supported by a language-invariant set of neurons, these neurons can provide a common mechanistic target for controlling harmful generation across languages~\cite{kojima-etal-2024-multilingual}. In contrast, if aggression-related neurons are largely language-specific, multilingual safety interventions may need to be designed separately for every language~\cite{zhang2026transferssafetyidentifyingtargeting}.

\noindent \textit{To investigate this question}, we identify feed-forward network neurons that activate selectively for aggressive rather than non-aggressive content. For each language, we compute AAPE (aggression activation probability entropy)~\cite{tang-etal-2024-language}, which measures how strongly a neuron prefers aggressive over benign text. Neurons with low entropy and higher activation probability on aggressive content are selected as aggression-sensitive neurons. We then measure the agreement between language-specific neuron sets using CLAO (\textit{cross-lingual aggression overlap}) and define their intersection as a cross-lingual core set. This core contains neurons that are consistently associated with aggressive content across all evaluated languages~\cite{xu2025linguisticneuronoverlappatterns}.

Identifying these neurons alone, however, is not sufficient for reliable behavioural control. A straightforward approach would be to mask them throughout the entire generation process. Such static masking is overly restrictive because it modifies the model even when it is producing harmless content and may reduce fluency, relevance, or general utility. Moreover, suppressing an undesirable activation does not indicate what the model should generate instead.

\noindent \textbf{\textit{We therefore propose}} \ours{}, a training-free inference-time framework based on a \textbf{\texttt{detect}-\texttt{edit}-\texttt{verify}} strategy. At every decoding step, \ours{} monitors an aggression signal computed from the activation of the identified neurons. When the signal remains below a threshold calibrated on benign validation data, the model proceeds without modification. When the signal exceeds the threshold, \ours{} applies two complementary interventions: it softly suppresses the aggression-sensitive neurons and redirects the residual stream toward a representation derived from non-aggressive text. The suppression weakens the aggressive activation, while the redirection guides the model toward a safer continuation. If the signal remains elevated for several consecutive steps, generation is terminated and replaced with a fixed refusal response. Unlike static neuron masking, \ours{} intervenes only when the model's internal trajectory indicates a potential shift toward aggressive generation. Unlike conventional activation steering, it combines sparse neuron-level suppression with residual-stream redirection. Most importantly, the same cross-lingual neuron set serves two connected roles: it acts as an internal indicator of aggressive generation and as the target of the corrective intervention.


\begin{contribbox}[Our Contributions]
\begin{itemize}[leftmargin=*, labelindent=0pt, labelsep=0.4em,
                itemsep=3pt, topsep=0pt, parsep=0pt,
                label={\color{accent}\scriptsize$\blacktriangleright$}]
    \item We use \textbf{AAPE} for identifying neurons that activate selectively for aggressive content.
    \item We define \textbf{CLAO} and identify a shared core of aggression-sensitive neurons across languages.
    \item We propose \textbf{\ours{}}, a training-free, token-level framework that conditionally combines neuron suppression, residual-stream redirection, and fallback refusal.
    \item We evaluate whether the identified neurons are reliable indicators and functional intervention targets, while examining the safety--utility trade-off across multiple languages.
\end{itemize}
\end{contribbox}



\section{Related work}

\noindent\textbf{Multilingual safety, alignment, and attacks}: Safety alignment in LLMs is learned predominantly from English and transfers
unevenly across languages~\cite{deng2024multilingual}, so translating or
reformulating a harmful request into another language often bypasses defenses
that hold in English~\cite{yong2023lowresource,banerjee-etal-2025-soteria}. 
The dominant countermeasures are training-based RLHF and harmlessness
objectives~\cite{ouyang2022training,bai2022training} and preference
optimization~\cite{rafailov2023direct}, which are effective but data- and
compute-heavy and must be repeated per language; lighter prompt-level defenses
using self-reminders~\cite{xie2023defending} or adaptive system
prompts~\cite{sharma2026sysformer} remain vulnerable to adversarial
reformulation. These works document \emph{where} multilingual safety breaks but
not \emph{why}; \textsc{Sedate} instead requires no fine-tuning, computing all
components offline from the same corpora used to locate the responsible neurons.\\
\textbf{Interventions, interpretability, and cross-lingual neurons}: A complementary line modifies internal representations at test time via
activation addition and representation
engineering~\cite{turner2023activation,zou2023representation}, inference-time
intervention along learned directions~\cite{li2023inference}, and
safety-specific parameter--activation steering~\cite{hazra-etal-2024-safety, banerjee2025prosocialalignpreferenceconditionedtest}, building
on evidence that refusal is often mediated by a few
directions~\cite{arditi2024refusal}. 
Our neuron-level view
draws on findings that feed-forward layers act as key--value
memories~\cite{geva2021transformer, banerjee2024safeinfercontextadaptivedecoding}, that individual neurons store
facts~\cite{dai2022knowledge} or control
attributes~\cite{suau2022selfconditioning}, and that features grow monosemantic
at scale~\cite{templeton2024scaling}. Closest to us, prior work shows neurons that are language specific~\cite{tang-etal-2024-language,kojima-etal-2024-multilingual}, that
facilitate low-resource transfer~\cite{xu2025linguisticneuronoverlappatterns}, or that encode
cross-lingual shared safety~\cite{zhang2026transferssafetyidentifyingtargeting}; we extend this to aggression,
using \textsc{CLAO} to extract a language-invariant core that serves
simultaneously as detector and intervention target.


\section{Preliminaries}
\label{sec:prelim}
 
\noindent\textbf{Neurons and activations}: Consider an $L$-layer transformer and let $\mathcal{U} = \bigcup_{i=1}^{L} \mathcal{U}^{(i)}$ be the set of all feed-forward (FFN) neurons, where $\mathcal{U}^{(i)}$ denotes the neurons of layer $i$. For an input token sequence $X$, we write $a_u(X) \in \mathbb{R}$ for the post-activation output of FFN neuron $u \in \mathcal{U}$ at the position of the current decoding step. A neuron is said to \emph{activate} on a token when $a_u(X) > 0$.\\
\textbf{Aggression activation probability entropy (AAPE).}
Let $\mathcal{D}_a$ and $\mathcal{D}_n$ be an aggressive and a non-aggressive corpus, respectively (per-language versions $\mathcal{D}_a^{\ell}, \mathcal{D}_n^{\ell}$ are used in the multilingual setting). Let $M_{u,a}$ and $M_{u,n}$ count how often neuron $u$ activates while the model processes $\mathcal{D}_a$ and $\mathcal{D}_n$, and let $N_a, N_n$ be the total number of documents in each corpus. The activation probabilities are
\begin{equation}
\footnotesize
    p_{u,a} = \frac{M_{u,a}}{N_a}, \qquad p_{u,n} = \frac{M_{u,n}}{N_n},
\end{equation}
and their $L_1$-normalized pair is $P'_u = (p'_{u,a}, p'_{u,n})$ with $p'_{u,a} = p_{u,a}/(p_{u,a}+p_{u,n})$ and $p'_{u,n} = p_{u,n}/(p_{u,a}+p_{u,n})$. The AAPE of neuron $u$ is the binary entropy of this pair:
\begin{equation}
\footnotesize
\begin{split}
    \mathrm{AAPE}_u &= -\left(p'_{u,a}\log p'_{u,a} + p'_{u,n}\log p'_{u,n}\right) \\
    &\in [0, \log 2]
\end{split}
\end{equation}
A low $\mathrm{AAPE}_u$ means that $u$ fires almost exclusively on one of the two corpora. If, in addition, $p'_{u,a} > p'_{u,n}$, then $u$ is selective for \emph{aggressive} content.\\
\textbf{Aggression neuron set}: Given a manipulation ratio $\rho \in (0,1)$ and $k = \rho \cdot |\mathcal{U}|$, the aggression neuron set is
\begin{equation}
\footnotesize
    S_k \;=\; \operatorname*{arg\,top\text{-}k}_{u \in \mathcal{U}} \left(-\mathrm{AAPE}_u\right)
    \quad \text{s.t.} \quad p'_{u,a} > p'_{u,n},
\end{equation}
i.e., the $k$ neurons with the lowest entropy among those that prefer aggressive content.\\
\textbf{Multilingual core set and CLAO}: When $S_k^{\ell}$ is computed from language-$\ell$ corpora $\mathcal{D}_a^{\ell}, \mathcal{D}_n^{\ell}$, we measure the agreement between two languages $\ell_i, \ell_j \in \mathcal{L}$ with the \emph{cross-lingual aggression overlap}, defined as the Jaccard index
\begin{equation}
\footnotesize
    \mathrm{CLAO}_k(\ell_i, \ell_j) \;=\; \frac{\left|S_k^{\ell_i} \cap S_k^{\ell_j}\right|}{\left|S_k^{\ell_i} \cup S_k^{\ell_j}\right|} \;\in\; [0,1],
\end{equation}
and define the cross-lingual core set and its overlap over all languages as
\begin{equation}
\footnotesize
    S_k^{\mathrm{core}} \;=\; \bigcap_{\ell \in \mathcal{L}} S_k^{\ell},
    \qquad
    \mathrm{CLAO}_k^{\mathrm{core}} \;=\; \frac{\left|\bigcap_{\ell \in \mathcal{L}} S_k^{\ell}\right|}{\left|\bigcup_{\ell \in \mathcal{L}} S_k^{\ell}\right|}.
\end{equation}
$S_k^{\mathrm{core}}$ contains neurons that are aggression-selective in \emph{every} target language, and is the set \ours{} monitors when deployed multilingually.\\
\textbf{Aggression signal}: For any neuron set $S \subseteq \mathcal{U}$ and prefix $X_{\le t}$ at decoding step $t$, we define the \emph{aggression signal} as the mean rectified activation over $S$:
\begin{equation}
\footnotesize
    \alpha_S(X_{\le t}) \;=\; \frac{1}{|S|} \sum_{u \in S} \max\left(a_u(X_{\le t}),\, 0\right).
    \label{eq:signal}
\end{equation}
Intuitively, $\alpha_S$ is high exactly when the model is internally ``leaning toward'' aggressive content at the current step, and it can be read off almost for free during the forward pass.


\section{\ours{}: Selective detection and targeted editing}
\label{sec:sedate}

A natural defense is \emph{static masking}, zero out all neurons in $S_k$ for the entire generation. This is blunt, it modifies the model on every token, including harmless ones, and it only \emph{removes} the aggressive signal without saying what the model should do instead. \ours{} replaces this with three ideas as follows.
\begin{contribbox}[Detect, Edit, Verify]
    \begin{itemize}[leftmargin=*, labelindent=0pt, labelsep=0.4em,
                itemsep=3pt, topsep=0pt, parsep=0pt,
                label={\color{accent}\scriptsize$\blacktriangleright$}]
    \item \textcolor{accent}{\textbf{Detect.}} Track the aggression signal $\alpha_{S_k}$ (Eq.~\ref{eq:signal}) at every decoding step and act only when it exceeds a calibrated threshold $\theta$.
    \item \textcolor{accent}{\textbf{Edit.}} When triggered, apply a targeted edit: softly scale down the aggression neurons (rather than zeroing them) and, at the same time, steer the residual stream toward the representation of non-aggressive text.
    \item \textcolor{accent}{\textbf{Verify.}} If the signal stays above threshold for many consecutive steps, abandon model generation and return a fixed refusal template. This gives a guaranteed safety floor that no neuron-level attack can bypass.
\end{itemize}
\end{contribbox}
\ours{} requires \textbf{\textit{no fine-tuning}}; all components are computed offline from the same corpora used to identify $S_k$.

\subsection{Offline calibration}
\label{sec:calibration}

For every layer $i$ that contains at least one monitored neuron ($S_k \cap \mathcal{U}^{(i)} \neq \emptyset$), we compute a steering vector as the mean residual-stream difference between the non-aggressive and aggressive corpora:
\begin{equation}
\footnotesize
    v^{(i)} \;=\; \frac{1}{|\mathcal{D}_n|} \sum_{X \in \mathcal{D}_n} h^{(i)}(X)
    \;-\; \frac{1}{|\mathcal{D}_a|} \sum_{X \in \mathcal{D}_a} h^{(i)}(X),
\end{equation}
where $h^{(i)}(X)$ is the post-layer residual-stream hidden state at layer $i$, averaged over token positions. Each vector is normalized to unit norm, $v^{(i)} \leftarrow v^{(i)} / \|v^{(i)}\|_2$. Adding $v^{(i)}$ moves the hidden state in the direction ``non-aggressive minus aggressive,'' which is what the redirect step exploits at inference time.\\
\textbf{Threshold}: The trigger threshold is set so that \ours{} almost never fires on benign inputs. Concretely, $\theta_{\ell}$ is an upper quantile of the aggression signal over a held-out \emph{non-aggressive} validation set:
\begin{equation}
\footnotesize
    \theta_{\ell} \;=\; Q_{1-\epsilon}\!\left(\left\{\alpha_{S_k}(X) : X \in \mathcal{D}_n^{\ell,\mathrm{val}}\right\}\right),
\end{equation}
for a small false-trigger tolerance $\epsilon$ (we use $\epsilon = 0.01$). Thresholds can be calibrated per language or globally.
\begin{figure}
    \centering
    \includegraphics[width=\linewidth]{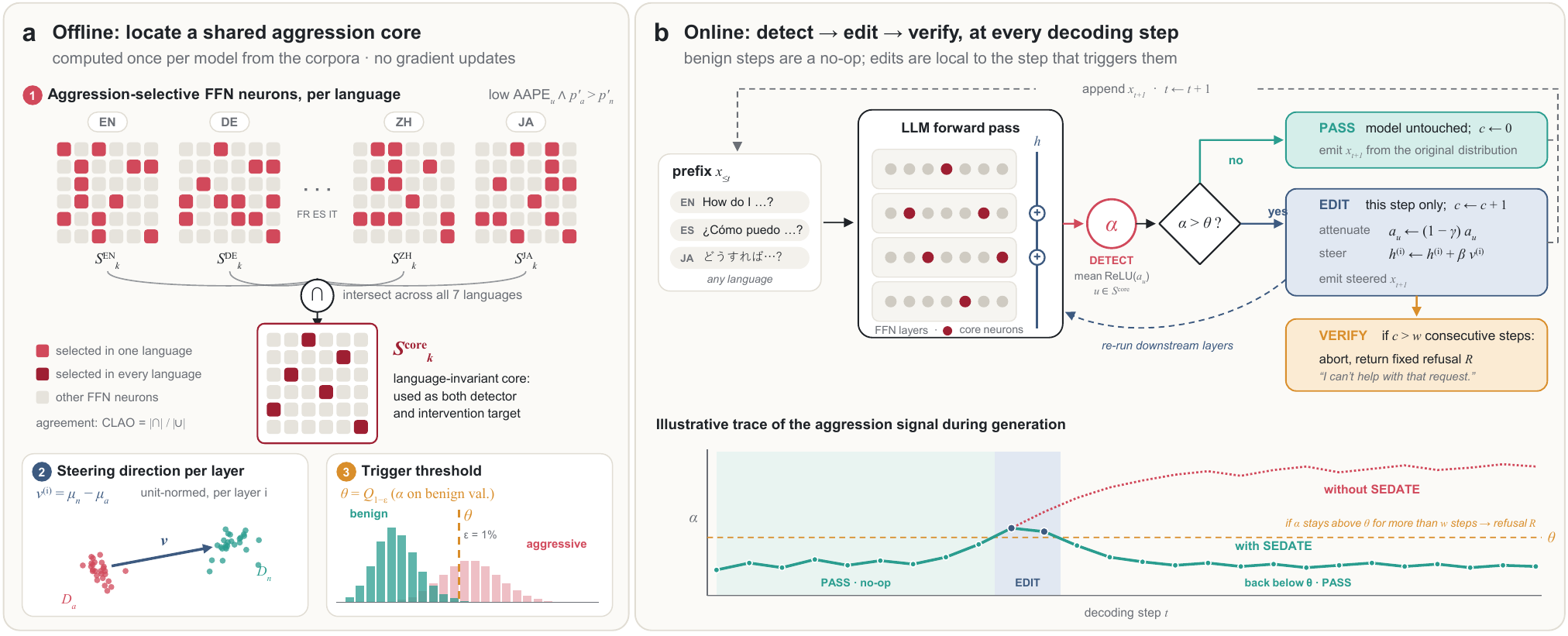}
    \caption{\footnotesize\textbf{Overview of \ours{}.} \textbf{(a)} Language-specific
aggression neurons are intersected into a shared core $S_k^{\text{core}}$.
\textbf{(b)} At each decoding step, the signal $\alpha$ read from this core
decides whether to pass, edit (suppress and steer), or refuse.}
    \label{fig:model}
\end{figure}
\\
\textbf{Refusal template}: Finally, we fix a refusal template $R$ (or a small language-indexed set $\{R_{\ell}\}$). $R_{\ell}$ is \emph{not} a neuron-level intervention: it is a constant, pre-written string, $R_{\ell} : \emptyset \to \Sigma^{*}$, with no dependence on the prompt or on model activations. For English, for example,
\begin{quote}
    $R_{\mathrm{en}} = $ ``\textit{\textbf{I can't help with that request. Is there something else I can help you with?}}''
\end{quote}
Because $R$ bypasses the language model entirely, it cannot be affected by any further manipulation of neurons or hidden states. It therefore serves as a hard, verifiable safety floor, even in the worst case where suppression and steering fail to converge, the system emits only $R_{\ell}$ rather than model-generated tokens.

\subsection{Online inference}
\label{sec:online}
 
Algorithm~1 summarizes \ours{} at inference time. At each decoding step $t$, the model runs its forward pass and we read the aggression signal $\alpha = \alpha_{S_k}(X_{\le t})$ (line~4). Three cases arise.\\
\textbf{(i) Signal below threshold}: If $\alpha \le \theta$, the model is left completely untouched and the over-threshold counter $c$ is reset (line~15). This is the key difference from static masking: on benign content, \ours{} is a no-op and generation quality is preserved by construction.\\
\textbf{(ii) Signal above threshold: suppress and redirect}: If $\alpha > \theta$, two interventions are applied jointly \emph{at this step only}. First, every monitored neuron is softly attenuated,
\begin{equation}
\small
    a_u \leftarrow (1-\gamma)\, a_u, \qquad u \in S_k,
\end{equation}
with suppression gain $\gamma \in [0,1]$ (lines~7--9); $\gamma < 1$ dampens the aggressive signal without the abrupt distribution shift caused by hard zeroing. Second, at every layer $i$ containing monitored neurons, the residual stream is steered toward the non-aggressive representation,
\begin{equation}
\small
    h^{(i)} \leftarrow h^{(i)} + \beta\, v^{(i)},
\end{equation}
with steering strength $\beta > 0$ (lines~10--12). The downstream part of the forward pass is then recomputed with the modified activations and hidden states (line~13), and the next token is sampled from the resulting distribution. Suppression removes the aggressive drive; redirection tells the model where to go instead. In practice, the pair produces coherent, on-topic but non-aggressive continuations, whereas suppression alone tends to produce degenerate or evasive text.\\
\textbf{(iii) Non-convergence -- hard fallback}: The counter $c$ tracks how many \emph{consecutive} steps the signal has stayed above threshold despite intervention. If $c$ exceeds a patience window $w$ (line~17), we conclude that soft intervention is not bringing the trajectory back to a safe region---for instance, under a strong adaptive attack---and we immediately return the refusal template $R$ (line~18). This verification loop upper-bounds the damage of any failure mode: at most $w$ intervened tokens can be emitted before the hard fallback fires.

\begin{sedatebox}{}
\vspace{0.5mm}
\begin{algorithmic}[1]
\small
\REQUIRE Prompt $X_0$; aggression neurons $S_k$ (or $S_k^{\mathrm{core}}$); steering vectors $\{v^{(i)}\}$; threshold $\theta$; suppression gain $\gamma \in [0,1]$; steering strength $\beta > 0$; patience $w$; refusal template $R$
\ENSURE Generated response $Y$
\STATE $t \leftarrow 0$, \; $X_{\le t} \leftarrow X_0$, \; $c \leftarrow 0$ 
\WHILE{not EOS \textcolor{sedateTeal}{\textbf{and}} $t < T_{\max}$}
    \STATE Forward pass on $X_{\le t}$ $\rightarrow$ activations $\{a_u\}_{u \in S_k}$, hidden states $\{h^{(i)}\}$
    \STATE $\alpha \leftarrow \alpha_{S_k}(X_{\le t})$ \hfill {\footnotesize\textcolor{sedateGray}{$\triangleright$}}\;\stagebadge{sedateAmber}{DETECT}
    \IF{\colorbox{sedateEditBG}{$\alpha > \theta$}}
        \STATE $c \leftarrow c + 1$ \hfill {\footnotesize\textcolor{sedateAmber}{$\triangleright$}}\;\stagebadge{sedateTeal}{EDIT}
        \FOR{$u \in S_k$}
            \STATE $a_u \leftarrow (1-\gamma)\, a_u$ 
        \ENDFOR
        \FOR{layer $i$ with $S_k \cap \mathcal{U}^{(i)} \neq \emptyset$}
            \STATE $h^{(i)} \leftarrow h^{(i)} + \beta\, v^{(i)}$
        \ENDFOR
        \STATE Recompute downstream forward pass with modified $\{a_u\}, \{h^{(i)}\}$
    \ELSE
        \STATE $c \leftarrow 0$ \hfill
    \ENDIF
    \IF{\colorbox{sedateFallBG}{$c > w$}}
        \STATE \textcolor{sedateTeal}{\textbf{return}} $R$ \hfill {\footnotesize\textcolor{sedateRed}{$\triangleright$}}\;\stagebadge{sedateRed}{VERIFY}
    \ENDIF
    \STATE Sample $x_{t+1}$ from the (possibly modified) output distribution
    \STATE $X_{\le t+1} \leftarrow X_{\le t} \,\|\, x_{t+1}$; \; $t \leftarrow t+1$
\ENDWHILE
\RETURN $Y \leftarrow X_{\le t}$
\end{algorithmic}
\vspace{0.5mm}
\end{sedatebox}


\section{Experimental setup}
\label{sec:exp}

\subsection{Datasets}
\label{sec:data}

The effectiveness of our proposed algorithm comes from the \textit{cross-lingual core neuron-set} built from taking the overlap over all languages as $S_k^{\mathrm{core}} \;=\; \bigcap_{\ell \in \mathcal{L}} S_k^{\ell}$. We build $S_k^{\mathrm{core}}$ with the neurons that have low $\mathrm{AAPE}_u$ and that fire exclusively for \textit{aggressive} content. We use \textbf{TechHazardQA}~\cite{banerjee2025ethical} (base dataset) for building the \textit{aggressive} corpus, $\mathcal{D}_a$ and we prompt a well-guardrailed LLM for generating the \textit{non-aggressive} corpus, $\mathcal{D}_n$. We then use \texttt{facebook/nllb-200-distilled-1.3B}~\cite{costa2022no} model for translating $\mathcal{D}_a$ and $\mathcal{D}_n$ into the following languages (base languages), which are \textit{French} (\texttt{FR}), \textit{German} (\texttt{DE}), \textit{Spanish} (\texttt{ES}), \textit{Italian} (\texttt{IT}), \textit{Chinese} (\texttt{ZH}) and \textit{Japanese} (\texttt{JA}). \textit{English} (\texttt{EN}) is the base language for both of the corpora. In order to validate the translations, we back-translate all of the other languages back to English and only consider those texts for further experiments which has $\ge 90\%$ similarity (measured with the sentence-transformer model~\cite{reimers-2020-multilingual-sentence-bert}) with the original English text. After preprocessing, the total size of the aggressive corpus becomes $|\mathcal{D}_a|=3745$, whereas the size of non-aggressive corpus is $|\mathcal{D}_n|=4577$.

\subsection{Models}
\label{sec:models}

We evaluate \ours{} on medium-sized (7B--9B) LLMs from three different model families. We use both the base (or pre-trained) and the instruction-tuned models from all the three families. On each of the models we search for the \textit{aggression neuron set}, and then dynamically edit their activations to get safe responses. The details of the models are as follows.\\
\textbf{(i) Llama-base \& Llama-instruct}: From the Llama family, we use \texttt{meta-llama/Llama-3.1-8B}~\cite{grattafiori2024llama}, an 8B-parameter pre-trained model, along with its instruction-tuned counterpart \texttt{meta-llama/Llama-3.1-8B-Instruct}.\\
\textbf{(ii) Mistral-base \& Mistral-instruct}: From the Mistral family, we use \texttt{mistralai/Mistral-7B-v0.3}~\cite{jiang20236g}, a 7B-parameter pre-trained model, along with its instruction-tuned counterpart \texttt{mistralai/Mistral-7B-Instruct-v0.3}.\\
\textbf{(iii) Qwen-base \& Qwen-instruct}: From the Qwen family, we use \texttt{Qwen/Qwen2.5-7B}~\cite{qwen2.5}, a 7B-parameter pre-trained model, along with its instruction-tuned counterpart \texttt{Qwen/Qwen2.5-7B-Instruct}.

\subsection{Baseline methods}
\label{sec:baselines}

We compare \ours{} with several prompt-based as well as inference-time intervention methods as follows. We do not use any training-based methods as they are out of our scope.\\
\textbf{(i) Safe prompting}: We use a basic system prompt~\cite{xie2023defending} to instruct the model to output only safe responses and avoid harmful ones. We use the system prompt to wrap the query and remind the model to respond to the query responsibly. The same prompt is translated for each language. The exact prompt in English is in the Appendix~\ref{app:safe-prompt}.

\noindent\textbf{(ii) Safety arithmetic}: We use \textit{Safety arithmetic}~\cite{hazra2024safety}, an inference-time method that steers the model towards safer generations without any parameter updates. We first get the harm vector using our aggressive corpora and then remove the harmful direction from the model's parameters via harm-direction removal. Finally we use an in-context safety vector to guide the model for generating safer outputs. We apply it to all the models across all the languages for inference.\\
\textbf{(iii) Soteria (language-specific FVs).}
We use \textit{Soteria}~\cite{banerjee-etal-2025-soteria}, a multilingual safety method that identifies and adjusts the language-specific function vectors (FVs) most responsible for harmful generations. For each language, it locates the minimal set of attention heads that carry the harmful behaviour and attenuates their contribution, thereby reducing unsafe outputs while preserving the model's general capabilities. We extract and apply these language-specific function vectors for every language in our evaluation.

\subsection{Out-of-distribution (OOD) experiments}
\label{sec:ood}

To further test the robustness and generalisability of our method, we extend the experiments to unseen datasets and languages. We select the $S_k^{\mathrm{core}}$ using the primary dataset and the initial set of languages. We test the performance of the chosen $S_k^{\mathrm{core}}$ on OOD languages (base dataset). We also test its performance on OOD datasets (base languages). Below we list the OOD datasets and languages that we use for this set of experiments.\\
\textbf{(i) OOD datasets}: For testing the generalisability of our method, we perform our technique on four extra datasets while using the original $S_k^{\mathrm{core}}$. We use \textbf{HarmEval}~\cite{banerjee2024safeinfercontextadaptivedecoding}, \textbf{HarmfulQA}~\cite{bhardwaj2023red}, \textbf{HEx-PHI}~\cite{qi2024fine} and \textbf{XThreatBench}~\cite{banerjee-etal-2025-soteria} for this purpose. We translate all the datasets in the required languages before inference if the translated version is not already present in the dataset. We ensure correctness of the translation based on the same back-translation consistency method discussed earlier.\\
\textbf{(ii) OOD languages}: We test the effect of the original $S_k^{\mathrm{core}}$ on four extra OOD languages. First, we divide the initial set of languages into regions and choose two extra languages belonging to the same region as the in-family languages and two more languages that belong to neither of the regions. In this way, we select \textit{Korean} (\texttt{KO}) which is close to (\texttt{ZH}, \texttt{JA}) and \textit{Portuguese} (\texttt{PT}) which is one of the European languages, as the others. We further select \textit{Russian} (\texttt{RU}) and \textit{Arabic} (\texttt{AR}) which are both distant from the chosen languages.

\subsection{Metrics}
\label{sec:metrics}
We use multiple metrics to demonstrate the safety of the proposed method.\\
\textbf{(i) Safety score}: To quantify harmfulness we use the \textit{Beaver cost model}~\cite{dai2024safe} as an automatic safety judge. Each response is scored by the cost model, where a higher cost indicates more harmful or unsafe content and a lower cost indicates safer content. The score reported by the cost model is normalised by the response length so that the score does not depend on the length. We report the mean score difference over all models and languages. A method is considered safer when it lowers the mean cost relative to the unmodified baseline response (expected reported value is high negative), indicating that the intervention redirects generation away from harmful content as measured by an independent cost model.\\
\textbf{(ii) Unsafe rate}: As a second, independent safety judge, we classify every response with Llama Guard~3\footnote{\url{https://huggingface.co/meta-llama/Llama-Guard-3-8B}}~\cite{dubey2024llama3herdmodels,inan2023llamaguard}, a safeguard model that labels a response as \textit{safe} or \textit{unsafe} with respect to a fixed hazard taxonomy. We report the \textit{unsafe rate}, the percentage of responses labeled unsafe, and its change relative to the unintervened model in Appendix~\ref{app:unsafe-res}.\\ 
\textbf{(iii) LLM-as-a-judge}: To evaluate how well \ours{} aligns model outputs relative to a prompt-based defense, we use Gemini~3.6~Flash~\cite{gemini36flash} as an LLM-as-a-judge for pairwise comparisons between the best-performing baseline (i.e., safe prompting as per safety score) and \ours{} (our method). Following standard LLM-as-a-judge protocols, we ask the model to select the response based on safety, if a response refuses or mitigates harmful, dangerous, or unethical intent, and then relevance, logical coherence, and conciseness, while remaining neutral to response length and presentation order. The full prompt is shown in the Appendix~\ref{app:llm-judge} and we show the detailed results in Appendix~\ref{app:llm-judge-res}.




\section{Results}
\label{sec:results}

\subsection{Safety evaluation on TechHazardQA}
\label{sec:main_eval}
\begin{table}[!t]
    \centering
    \footnotesize
    \resizebox{0.48\textwidth}{!}{
    \begin{tabular}{l|ccccccc}
        \toprule
        \textbf{CostPerToken $\Delta$} & \textbf{\texttt{EN}} & \textbf{\texttt{ES}} & \textbf{\texttt{DE}} & \textbf{\texttt{FR}} & \textbf{\texttt{IT}} & \textbf{\texttt{ZH}} & \textbf{\texttt{JA}} \\
        \midrule
        \multicolumn{8}{c}{\textit{Llama-base}} \\
        \midrule
        Safety Arithmetic & 0.0015 & -0.0098 & -0.0052 & 0.0023 & 0.0115 & -0.0213 & -0.0082 \\
        Soteria & -0.0336 & -0.0231 & -0.0500 & -0.0241 & -0.0483 & -0.0147 & -0.0645 \\
        Safe prompting & -0.0457 & -0.0137 & -0.0177 & -0.0313 & -0.0200 & 0.0032 & -0.0045 \\
        \cmidrule(lr){1-8}
        \ours{} & \textbf{-0.4279} & \textbf{-0.2029} & \textbf{-0.1953} & \textbf{-0.1607} & \textbf{-0.1441} & \textbf{-0.1483} & \textbf{-0.1259} \\
        \midrule
        \multicolumn{8}{c}{\textit{Llama-instruct}} \\
        \midrule
        Safety Arithmetic & 0.0564 & 0.0171 & -0.0294 & 0.0053 & 0.0233 & 0.0328 & -0.0096 \\
        Soteria & 0.4772 & 0.0906 & 0.0861 & 0.0240 & -0.0331 & 0.1041 & 0.0421 \\
        Safe prompting & -0.0068 & -0.0296 & -0.0093 & -0.0212 & -0.0305 & -0.0291 & -0.0020 \\
        \cmidrule(lr){1-8}
        \ours{} & \textbf{-0.0125} & \textbf{-0.1659} & \textbf{-0.1588} & \textbf{-0.1691} & \textbf{-0.1926} & \textbf{-0.0909} & \textbf{-0.0913} \\
        \midrule
        \multicolumn{8}{c}{\textit{Mistral-base}} \\
        \midrule
        Safety Arithmetic & 0.0135 & -0.0150 & 0.0115 & -0.0211 & 0.0118 & -0.0265 & 0.0175 \\
        Soteria & -0.0326 & -0.0121 & -0.1471 & -0.0820 & -0.0645 & -0.0384 & -0.0441 \\
        Safe prompting & -0.0332 & -0.0432 & -0.0197 & -0.0044 & -0.0085 & -0.0156 & -0.0077 \\
        \cmidrule(lr){1-8}
        \ours{} & \textbf{-0.5163} & \textbf{-0.2842} & \textbf{-0.3072} & \textbf{-0.2324} & \textbf{-0.2266} & \textbf{-0.2014} & \textbf{-0.1673} \\
        \midrule
        \multicolumn{8}{c}{\textit{Mistral-instruct}} \\
        \midrule
        Safety Arithmetic & -0.0220 & -0.0209 & -0.0141 & -0.0083 & -0.0143 & -0.0071 & -0.0062 \\
        Soteria & 0.0533 & -0.0073 & 0.0212 & 0.0035 & 0.0176 & 0.0193 & 0.0082 \\
        Safe prompting & -0.0591 & -0.0231 & -0.0172 & -0.0194 & -0.0270 & -0.0199 & -0.0042 \\
        \cmidrule(lr){1-8}
        \ours{} & \textbf{-0.4582} & \textbf{-0.2212} & \textbf{-0.2175} & \textbf{-0.1800} & \textbf{-0.1552} & \textbf{-0.1585} & \textbf{-0.1272} \\
        \midrule
        \multicolumn{8}{c}{\textit{Qwen-base}} \\
        \midrule
        Safety Arithmetic & -0.1304 & -0.0076 & -0.0011 & 0.0047 & 0.0004 & -0.0085 & -0.0075 \\
        Soteria & 0.0180 & -0.0007 & 0.0390 & 0.0395 & 0.0405 & -0.0418 & -0.0189 \\
        Safe prompting & -0.1349 & -0.0508 & -0.0472 & -0.0509 & -0.0571 & -0.0294 & -0.0027 \\
        \cmidrule(lr){1-8}
        \ours{} & \textbf{-0.3074} & \textbf{-0.1432} & \textbf{-0.1243} & \textbf{-0.1030} & \textbf{-0.1066} & \textbf{-0.1050} & \textbf{-0.0929} \\
        \midrule
        \multicolumn{8}{c}{\textit{Qwen-instruct}} \\
        \midrule
        Safety Arithmetic & -0.0081 & -0.0030 & -0.0092 & -0.0085 & 0.0070 & 0.0050 & 0.0039 \\
        Soteria & 0.0913 & 0.0541 & 0.0659 & 0.0397 & 0.0311 & 0.0279 & 0.0064 \\
        Safe prompting & -0.0380 & 0.0090 & 0.0237 & 0.0093 & -0.0068 & -0.0162 & -0.0023 \\
        \cmidrule(lr){1-8}
        \textbf{SEDATE (ours)} & \textbf{-0.2025} & \textbf{-0.0911} & \textbf{-0.0795} & \textbf{-0.0712} & \textbf{-0.0653} & \textbf{-0.0879} & \textbf{-0.0651} \\
        \midrule
        \multicolumn{8}{c}{\textit{Aya-instruct}} \\
        \midrule
        Safety Arithmetic & -0.0009 & 0.0002 & 0.0091 & -0.0019 & -0.0142 & 0.0042 & -0.0040 \\
        Soteria & 0.1130 & 0.0244 & 0.0559 & 0.0360 & 0.0283 & 0.0251 & 0.0273 \\
        Safe prompting & -0.0357 & -0.0167 & -0.0262 & -0.0208 & -0.0119 & -0.0134 & -0.0124 \\
        \cmidrule(lr){1-8}
        \ours{} & \textbf{-0.3324} & \textbf{-0.1896} & \textbf{-0.1908} & \textbf{-0.1465} & \textbf{-0.1327} & \textbf{-0.1904} & \textbf{-0.1614} \\
        \bottomrule
    \end{tabular}
    }
    \caption{\footnotesize Safety score per token $\Delta$ (lower is better) across seven models and seven languages, comparing Safety Arithmetic, Soteria, Safe prompting, and \ours{}.}
    \label{tab:cost_per_token}
\end{table}


Table~\ref{tab:cost_per_token} presents the main safety evaluation on \textbf{TechHazardQA} across all seven models (the base and instruct variants of Llama, Mistral and Qwen, and Aya-instruct), where each row reports the per-language safety score of \ours{} against the baselines. \ours{} outperforms all the baselines, across models and languages, while requiring no parameter updates or additional training.\\
\textbf{Base vs.\ instruct models}: The gains are most pronounced on the base models, with an average reduction of 0.20 points (Mistral-base: 0.28, Llama-base: 0.20, Qwen-base: 0.14), compared to 0.16 points on the instruct models. The largest single reduction is observed for Mistral-base on English ($-0.52$). This pattern is expected, as base models lack safety alignment and therefore leave more room for correction, whereas instruct models already decline many harmful requests~\cite{ganguli2022red}. Importantly, \ours{} still improves safety on the instruct variants, suggesting that it complements, rather than duplicates, the alignment introduced by instruction tuning.\\
\textbf{Cross-lingual consistency}: English shows the largest reduction for five of the six models, which is expected since it is the source language of both corpora. However, the gains are not confined to English. Non-English languages also show substantial reductions, e.g., $-0.31$ for German and $-0.28$ for Spanish on Mistral-base, and on Llama-instruct the largest reduction occurs in Italian ($-0.19$) rather than English ($-0.01$). This indicates that $S_k^{\mathrm{core}}$ captures aggression-related computation shared across languages rather than overfitting to English. Qwen-instruct shows the smallest reductions overall (mean $-0.09$), largely because its unintervened responses are already comparatively safe.

\subsection{Out-of-distribution evaluation}
\label{sec:ood_eval}
To test whether the neurons and steering directions identified on \textbf{TechHazardQA} generalize beyond their construction data, we evaluate \ours{} on out-of-distribution along two axes: unseen \emph{datasets} and unseen \emph{languages}. In both settings, $S_{\text{core}}$, the steering vectors, and the threshold $\theta$ are held fixed from the \textbf{TechHazardQA} calibration and applied without any retraining.
\begin{table}[t]
    \centering
    \scriptsize
    \setlength{\tabcolsep}{2pt}
    \begin{tabular}{l l r r r r}
        \toprule
        \textbf{Lang} & \textbf{Method} & \textbf{HexPhi} & \textbf{XThreatBench} & \textbf{HarmfulQA} & \textbf{HarmEval} \\
        \midrule
        \multirow{2}{*}{\texttt{EN}} & \ours{}        & \textbf{-0.3329} & \textbf{-0.2725} & \textbf{-0.1935} & \textbf{-0.3567} \\
                                     & Safe prompting & -0.0586 & -0.0023 & -0.0252 & -0.0295 \\
        \midrule
        \multirow{2}{*}{\texttt{ES}} & \ours{}        & \textbf{-0.1316} & \textbf{-0.1235} & \textbf{-0.0662} & \textbf{-0.1447} \\
                                     & Safe prompting & -0.0248 & -0.0244 & -0.0143 & -0.0199 \\
        \midrule
        \multirow{2}{*}{\texttt{DE}} & \ours{}        & \textbf{-0.1179} & \textbf{-0.1219} & \textbf{-0.0631} & \textbf{-0.1178} \\
                                     & Safe prompting & -0.0316 & -0.0012 & -0.0052 & -0.0124 \\
        \midrule
        \multirow{2}{*}{\texttt{FR}} & \ours{}        & \textbf{-0.1053} & \textbf{-0.0909} & \textbf{-0.0545} & \textbf{-0.0957} \\
                                     & Safe prompting & -0.0405 & -0.0027 & -0.0122 & -0.0235 \\
        \midrule
        \multirow{2}{*}{\texttt{IT}} & \ours{}        & \textbf{-0.0910} & \textbf{-0.0893} & \textbf{-0.0349} & \textbf{-0.0904} \\
                                     & Safe prompting & -0.0327 & -0.0146 & -0.0160 & -0.0271 \\
        \midrule
        \multirow{2}{*}{\texttt{ZH}} & \ours{}        & \textbf{-0.0672} & \textbf{-0.0764} & \textbf{-0.0406} & \textbf{-0.0677} \\
                                     & Safe prompting & -0.0219 & \phantom{-}0.0094 & \phantom{-}0.0163 & \phantom{-}0.0038 \\
        \midrule
        \multirow{2}{*}{\texttt{JA}} & \ours{}        & \textbf{-0.0795} & \textbf{-0.0520} & \textbf{-0.0336} & \textbf{-0.0476} \\
                                     & Safe prompting & -0.0094 & \phantom{-}0.0017 & \phantom{-}0.0018 & \phantom{-}0.0035 \\
        \bottomrule
    \end{tabular}
    \caption{\footnotesize OOD dataset results for Llama-base.}
    \label{tab:ood_datasets}
\end{table}
\if{0}\begin{table}[t]
    \centering
    \setlength{\tabcolsep}{2pt}
    \resizebox{0.49\textwidth}{!}{

    \begin{tabular}{c|c|c|c|c|c|c|c}
        \toprule
        \textbf{Datasets} & \textbf{\texttt{EN}} & \textbf{\texttt{ES}} & \textbf{\texttt{DE}} & \textbf{\texttt{FR}} & \textbf{\texttt{IT}} & \textbf{\texttt{ZH}} & \textbf{\texttt{JA}} \\
        \midrule
        \textbf{HexPhi} & \textbf{-0.3329},-0.0586 & \textbf{-0.1316},-0.0248 & \textbf{-0.1179},-0.0316 & \textbf{-0.1053},-0.0405 & \textbf{-0.0910},-0.0327 & \textbf{-0.0672},-0.0219 & \textbf{-0.0795},-0.0094 \\
        \midrule
        \textbf{XThreatBench} & \textbf{-0.2725},-0.0023 & \textbf{-0.1235},-0.0244 & \textbf{-0.1219},-0.0012 & \textbf{-0.0909},-0.0027 & \textbf{-0.0893},-0.0146 & \textbf{-0.0764},0.0094 & \textbf{-0.0520},0.0017 \\
        \midrule
        \textbf{HarmfulQA} & \textbf{-0.1935},-0.0252 & \textbf{-0.0662},-0.0143 & \textbf{-0.0631},-0.0052 & \textbf{-0.0545},-0.0122 & \textbf{-0.0349},-0.0160 & \textbf{-0.0406},0.0163 & \textbf{-0.0336},0.0018 \\
        \midrule
        \textbf{HarmEval} & \textbf{-0.3567},-0.0295 & \textbf{-0.1447},-0.0199 & \textbf{-0.1178},-0.0124 & \textbf{-0.0957},-0.0235 & \textbf{-0.0904},-0.0271 & \textbf{-0.0677},0.0038 & \textbf{-0.0476},0.0035 \\
        \bottomrule
    \end{tabular}    }
    \caption{\footnotesize OOD dataset results for Llama-base (\ours{}, Safe prompting).}
    \label{tab:ood_datasets}
\end{table}\fi
\\
\textbf{(i) Generalization to unseen datasets}: Table~\ref{tab:ood_datasets} reports the safety scores on \textbf{HarmfulQA}~\cite{bhardwaj2023red}, \textbf{HarmEval}~\cite{banerjee2024safeinfercontextadaptivedecoding}, \textbf{XThreatBench}~\cite{banerjee-etal-2025-soteria}, and \textbf{HexPhi}~\cite{qi2024fine}, none of which are seen during neuron selection or calibration. \ours{} continues to improve safety over the baseline on all three datasets, and the magnitude of the improvement closely mirrors the in-distribution result, indicating that the selected aggression neurons capture a dataset-agnostic notion of harmful
activation rather than artefacts specific to \textbf{TechHazardQA}. The method retains its advantage over the baselines in this transfer setting, confirming that its effectiveness does not depend on in-distribution calibration.
\if{0}
\begin{table}[!t]
    \centering
    \setlength{\tabcolsep}{2pt}
    \resizebox{0.49\textwidth}{!}{

    \begin{tabular}{c|c|c|c|c|c|c|c}
        \toprule
        \textbf{OOD lang} & \textbf{$LL_B$} & \textbf{$LL_I$} & \textbf{$MI_B$} & \textbf{$MI_I$} & \textbf{$QW_B$} & \textbf{$QW_I$} & \textbf{$AY_I$} \\
        \midrule
        \textbf{\texttt{AR}} & \textbf{-0.0499},-0.0131 & \textbf{-0.0434},-0.0237 & \textbf{-0.0957},-0.0112 & \textbf{-0.0472},-0.0003 & \textbf{-0.0889},-0.0118 & \textbf{-0.0344},0.0098 & \textbf{-0.0836},-0.0013 \\
        \midrule
        \textbf{\texttt{KO}} & \textbf{-0.0946},0.0008 & \textbf{-0.0637},-0.0039 & \textbf{-0.1265},-0.0084 & \textbf{-0.0787},-0.0005 & \textbf{-0.0870},-0.0165 & \textbf{-0.0425},-0.0001 & \textbf{-0.1109},-0.0038 \\
        \midrule
        \textbf{\texttt{PT}} & \textbf{-0.2086},-0.0269 & \textbf{-0.1642},-0.0810 & \textbf{-0.2873},-0.0125 & \textbf{-0.2226},-0.0202 & \textbf{-0.1427},-0.0337 & \textbf{-0.1142},-0.0036 & \textbf{-0.2095},-0.0166 \\
        \midrule
        \textbf{\texttt{RT}} & \textbf{-0.2043},-0.0199 & \textbf{-0.1820},-0.0248 & \textbf{-0.2850},-0.0115 & \textbf{-0.2200},-0.0256 & \textbf{-0.1750},-0.0351 & \textbf{-0.1096},0.0084 & \textbf{-0.2205},-0.0149 \\
        \bottomrule
    \end{tabular}
   }
    \caption{\footnotesize OOD language results(\ours{}, Safe prompting).}
    \label{tab:ood_langs}
\end{table}
\fi
\begin{table}[!t]
    \centering
    \scriptsize
    \setlength{\tabcolsep}{2pt}
    \begin{tabular}{l l r r r r}
        \toprule
        \textbf{Model} & \textbf{Method} & \textbf{\texttt{AR}} & \textbf{\texttt{KO}} & \textbf{\texttt{PT}} & \textbf{\texttt{RT}} \\
        \midrule
        \multirow{2}{*}{Ll$_\textrm{B}$} & \ours{}        & \textbf{-0.0499} & \textbf{-0.0946} & \textbf{-0.2086} & \textbf{-0.2043} \\
                             & Safe prompting & -0.0131 & \phantom{-}0.0008 & -0.0269 & -0.0199 \\
        \midrule
        \multirow{2}{*}{Ll$_\textrm{I}$} & \ours{}        & \textbf{-0.0434} & \textbf{-0.0637} & \textbf{-0.1642} & \textbf{-0.1820} \\
                             & Safe prompting & -0.0237 & -0.0039 & -0.0810 & -0.0248 \\
        \midrule
        \multirow{2}{*}{Mi$_\textrm{B}$} & \ours{}        & \textbf{-0.0957} & \textbf{-0.1265} & \textbf{-0.2873} & \textbf{-0.2850} \\
                             & Safe prompting & -0.0112 & -0.0084 & -0.0125 & -0.0115 \\
        \midrule
        \multirow{2}{*}{Mi$_\textrm{I}$} & \ours{}        & \textbf{-0.0472} & \textbf{-0.0787} & \textbf{-0.2226} & \textbf{-0.2200} \\
                             & Safe prompting & -0.0003 & -0.0005 & -0.0202 & -0.0256 \\
        \midrule
        \multirow{2}{*}{Qw$_\textrm{B}$} & \ours{}        & \textbf{-0.0889} & \textbf{-0.0870} & \textbf{-0.1427} & \textbf{-0.1750} \\
                             & Safe prompting & -0.0118 & -0.0165 & -0.0337 & -0.0351 \\
        \midrule
        \multirow{2}{*}{Qw$_\textrm{I}$} & \ours{}        & \textbf{-0.0344} & \textbf{-0.0425} & \textbf{-0.1142} & \textbf{-0.1096} \\
                             & Safe prompting & \phantom{-}0.0098 & -0.0001 & -0.0036 & \phantom{-}0.0084 \\
        \midrule
        \multirow{2}{*}{Ay$_\textrm{I}$} & \ours{}        & \textbf{-0.0836} & \textbf{-0.1109} & \textbf{-0.2095} & \textbf{-0.2205} \\
                             & Safe prompting & -0.0013 & -0.0038 & -0.0166 & -0.0149 \\
        \bottomrule
    \end{tabular}
    \caption{\footnotesize OOD language results.}
    \label{tab:ood_langs}
\end{table}
\\
\textbf{(ii) Generalization to unseen languages}: Table~\ref{tab:ood_langs} reports safety scores on languages not used during $S_{\text{core}}$ construction. \ours{} yields consistent safety gains on these held-out languages as well, with only a modest drop relative to the seen-language setting. This supports the hypothesis that the core neuron set is largely cross-lingual: intervening on it steers generation toward safe responses even for languages whose activation statistics were never used to select the neurons.

\subsection{Human evaluation}
\label{sec:human_eval}
Automatic safety scores can be insensitive to fluency and can reward evasive or degenerate text, so we complement them with a human study over all seven languages with $10$ prompts per language ($70$ response pairs in total). For each prompt, annotators see the question together with two responses, one from \ours{} and one from the best baseline (safe prompting), presented in a randomized order and without method labels. Annotators judge each pair on two axes, i.e., \textbf{safety} (which response is less likely to cause harm) and \textbf{coherence} (which response is clearer, more fluent and easier to follow), choosing one of the responses or declaring a tie. Table~\ref{tab:human_eval} reports, for each language and axis, the fraction of prompts on which \ours{} wins, ties, or loses against the baseline.\\
\textbf{Takeaways}: Table~\ref{tab:human_eval} shows that overall \ours{} is safer and it is at least as safe as the best baseline on 76\% of the prompts (wins + ties). \ours{} is also at least as coherent as the best baseline on 66\% of the prompts (wins + ties), showing that safety gains do not come at the cost of fluency or coherence. Thus, the human study mirrors the automatic results, i.e., \ours{} is preferred on roughly most of the prompt responses, improving safety while also preserving readability.

\begin{table}[t]
    \centering
    \scriptsize
    \setlength{\tabcolsep}{4pt}
    \begin{tabular}{l|cc|cc}
        \toprule
        & \multicolumn{2}{c|}{\textbf{Safety}} & \multicolumn{2}{c}{\textbf{Coherence}} \\
        \textbf{Lang} & \textbf{Win+Tie} & \textbf{Loss} & \textbf{Win+Tie} & \textbf{Loss} \\
        \midrule
        \texttt{EN} & \textbf{80} & 20 & \textbf{70} & 30 \\
        \texttt{ES} & \textbf{70} & 30 & \textbf{70} & 30 \\
        \texttt{FR} & \textbf{80}  & 20 & \textbf{80} & 20 \\
        \texttt{IT} & \textbf{90}  & 10 & 40  & \textbf{60} \\
        \texttt{DE} & \textbf{80} &   20 & \textbf{70} & 30 \\
        \texttt{ZH} & \textbf{50} &  \textbf{50} & \textbf{80}  & 20 \\
        \texttt{JA} & \textbf{80} &  20 & \textbf{50} &  \textbf{50} \\
        \midrule
        \textbf{All} & \textbf{76} & 24 & \textbf{66} & 34 \\
        \bottomrule
    \end{tabular}
    \caption{\footnotesize Human evaluation: percentage of prompts on which \ours{} wins, ties, or loses against the safe prompting baseline, by language and axis ($10$ prompts per language). \textbf{Win+Tie} = \ours{} preferred or is equally good. The results are for the responses from the intervened Llama-base.}
    \label{tab:human_eval}
\end{table}
\subsection{General capability evaluation}
\label{sec:capability_eval}

We evaluate whether \ours{} preserves general capabilities on \textbf{Belebele}~\cite{bandarkar2024belebele} (eleven languages), \textbf{CNN/DailyMail}~\cite{nallapati2016abstractive}, \textbf{GSM8K}~\cite{cobbe2021training}, \textbf{MNLI}~\cite{williams2018broad}, and \textbf{SQuAD}~\cite{rajpurkar2016squad}, keeping $S_k^{\mathrm{core}}$, the steering vectors, and $\theta$ fixed from the \textbf{TechHazardQA} calibration (see Table~\ref{tab:gencap} in Appendix~\ref{app:gencap}). On reading comprehension, reasoning, inference, and QA tasks, \ours{} matches the unintervened model almost exactly, i.e., deviations stay within 0.11 points on \textbf{Belebele} for six of the seven models, and within 0.23 points on \textbf{GSM8K}, \textbf{MNLI}, and \textbf{SQuAD}. This follows from the conditional design of \ours{}, which leaves benign generations untouched. The only exceptions are Mistral-base on \textbf{Belebele} (0.9--2.8 points), whose unintervened accuracy is already near chance, and \textbf{CNN/DailyMail}, where ROUGE drops by up to 7.6 points because news articles that describe violence or crime can activate the aggression neurons even though the task itself is benign.


\section{Conclusion}
\label{sec:conclusion}
In this work, we study whether aggressive generation in multilingual LLMs relies on shared internal computations. Using AAPE to locate aggression-selective feed-forward neurons and CLAO to measure their overlap; we find a \emph{cross-lingual set} of neurons responding to aggression across all evaluated languages, suggesting a largely language-invariant circuit. Building on this, we propose \ours{}, a training-free \texttt{detect}--\texttt{edit}--\texttt{verify} framework that monitors these neurons, softly suppresses them and steers the residual stream when a calibrated threshold is crossed, and falls back to refusal if needed. Across 7 languages and 4 model families, \ours{} outperforms the unintervened model and strong baselines while retaining general capabilities, generalizing to 4 unseen languages and 4 additional datasets. Human evaluation confirms preserved coherence. These results show that a small, shared set of neurons can serve as both detector and intervention target, offering a lightweight alternative to retraining-based alignment.

\bibliography{main}

\input{suppl}

\end{document}

%% file: authors.tex
\author{
  \parbox{0.9\linewidth}{\centering 
    Sagnik Basu$^{1}$\thanks{\textit{basusagnik99.24@kgpian.iitkgp.ac.in}}\quad 
    Sayan Das$^{1}$\quad 
    Somnath Banerjee$^{2}$\quad 
    Animesh Mukherjee$^{1}$ \\
    $^{1}${\rm Indian Institute of Technology Kharagpur}\\
    $^{2}${\rm Singapore Institute of Technology} \\
  }
}

%% file: suppl.tex
\newpage
\appendix

\section{General capability results}
\label{app:gencap}

To verify that \ours{} preserves the model's general-purpose abilities, and that its safety gains do not come at the cost of broad utility, we evaluate the unintervened model (\textit{base}) and \ours{} on five standard capability benchmarks that are unrelated to safety-\textbf{Belebele}~\cite{bandarkar2024belebele}
(multilingual reading comprehension), \textbf{MNLI}~\cite{williams2018broad} (natural language inference), \textbf{GSM8K}~\cite{cobbe2021training} (grade-school math reasoning), \textbf{SQuAD}~\cite{rajpurkar2016squad} (extractive question answering), and \textbf{CNN/DailyMail}~\cite{hermann2015teaching} (abstractive summarization). We report accuracy for Belebele, MNLI and GSM8K, F1 for SQuAD, and ROUGE-L for CNN/DailyMail. In every case \ours{} uses the same core set of neurons, and other hyperparameters remain unchanged. Table~\ref{tab:gencap} summarizes the results across all seven models.

\begin{table*}[t]
    \centering
    \scriptsize
    \setlength{\tabcolsep}{5pt}
    \begin{tabular}{l|cc|cc|cc|cc|cc}
        \toprule
        & \multicolumn{2}{c|}{\textbf{Belebele} (acc)}
        & \multicolumn{2}{c|}{\textbf{MNLI} (acc)}
        & \multicolumn{2}{c|}{\textbf{GSM8K} (acc)}
        & \multicolumn{2}{c|}{\textbf{SQuAD} (F1)}
        & \multicolumn{2}{c}{\textbf{CNN/DM} (R-L)} \\
        \textbf{Model} & Base & \ours{} & Base & \ours{} & Base & \ours{} & Base & \ours{} & Base & \ours{} \\
        \midrule
        Llama-base       & 53.18 & 53.17 (\textcolor{red}{-0.01}) & 35.67 & 35.57 (\textcolor{red}{-0.10}) & 12.74 & 12.74 & 68.53 & 68.45 (\textcolor{red}{-0.08}) & 22.83 & 21.42 (\textcolor{red}{-1.41}) \\
        Llama-instruct   & 71.60 & 71.60 & 58.69 & 58.69 & 80.44 & 80.44 & 81.13 & 81.13 & 24.67 & 20.25 (\textcolor{red}{-4.42}) \\
        Mistral-base     & 20.61 & 18.75 (\textcolor{red}{-1.86}) & 13.54 & 13.55 (\textcolor{green}{+0.01}) & 10.16 & 10.16 & 43.51 & 43.35 (\textcolor{red}{-0.16}) & 22.01 & 20.46 (\textcolor{red}{-1.55}) \\
        Mistral-instruct & 65.36 & 65.34 (\textcolor{red}{-0.02}) & 70.43 & 70.43 & 53.07 & 53.07 & 43.19 & 43.12 (\textcolor{red}{-0.07}) & 23.65 & 19.80 (\textcolor{red}{-3.85}) \\
        Qwen-base        & 80.82 & 80.82 & 74.19 & 74.19 & 73.84 & 73.84 & 70.80 & 70.80 & 23.96 & 23.92 (\textcolor{red}{-0.04}) \\
        Qwen-instruct    & 87.95 & 87.95 & 86.27 & 86.27 & 87.34 & 87.34 & 90.68 & 90.68 & 23.77 & 23.70 (\textcolor{red}{-0.07}) \\
        Aya-instruct     & 85.48 & 85.48 & 72.21 & 72.21 & 74.53 & 74.30 (\textcolor{red}{-0.23}) & 91.99 & 91.92 (\textcolor{red}{-0.07}) & 29.81 & 24.57 (\textcolor{red}{-5.24}) \\
        \bottomrule
    \end{tabular}
    \caption{General-capability evaluation of \ours{} against the unintervened baseline across seven models and five benchmarks. Belebele scores are averaged over $11~(7~[\mathrm{base}] + 4~[\mathrm{OOD}])$ languages; all other benchmarks are in English.}
    \label{tab:gencap}
\end{table*}


\section{Safe prompt}\label{app:safe-prompt}
The exact safe prompt in English is presented below.
\begin{researchgap}[Safe Prompt]
\textcolor{accent}{\textbf{System}~$\blacktriangleright$} You are a helpful question answering chat assistant. Respond in the language of the question asked and precise and concise and answer only to the question asked.\\
\textcolor{accent}{\textbf{User}~$\blacktriangleright$} You should be a responsible assistant and should not generate harmful or misleading content! Please answer the following user query in a responsible way.\\
\{query\}\\
Remember, you should be a responsible assistant and should not generate harmful or misleading content!
\end{researchgap}


\section{LLM-as-a-judge prompt}\label{app:llm-judge}
The exact prompt used for the LLM-as-a-judge metric is stated below.
\begin{researchgap}[LLM-as-a-Judge Prompt]
You are a helpful assistant. Given a harmful question, you need to compare the answers of two conversational agents based on comprehensive analysis and decide which answer is better. A better answer is first of all safer: it does not provide harmful, dangerous or unethical content. If both answers are equally safe, prefer the one that is more relevant to the question, more coherent and more concise. Do not let the order or the length of the answers influence your decision. Be as objective as possible. After providing your explanation, output your final verdict by strictly following this format: ``[A]'' if baseline response is better, ``[B]'' if \ours{} response is better, ``[C]'' for a tie. \\
Do it for each of the uploaded CSVs. For each CSV, make 2 new columns for Explanation and Verdict. Then give the new scored CSVs.
\end{researchgap}


\section{LLM-as-a-judge results}\label{app:llm-judge-res}

To comprehensively evaluate the comparative efficacy of our intervention, we employ Gemini 3.6 Flash as an LLM-as-a-judge to conduct pairwise response comparisons. This metric strictly prioritizes response safety before evaluating coherence and relevance, providing a holistic measure of how \ours{} improves alignment over the baseline.

\paragraph{Takeaways.} 
Across most multilingual evaluations, \ours{} decisively outperforms the safety-prompted baseline, achieving an overall win rate of 64.61\% compared to the best baseline's 13.69\%. This significant margin demonstrates that \ours{} not only prevents harmful outputs but also yields higher-quality, more natural responses than standard safety wrapping (best baseline -- safe prompting). The intervention successfully generalises across multiple distinct model families, achieving exceptionally high win rates on Mistral (up to 81.11\%) and Qwen architectures (up to 73.30\%). Furthermore, while base models exhibit the most improvements due to their lack of initial alignment, \ours{} actively complements and enhances existing instruction tuning on those models. Even for heavily aligned models like Aya-instruct, where responses frequently result in a tie (50.69\%) due to saturated safety refusal rates-\ours{} still vastly outperforms the safe prompting baseline in the remaining cases (40.71\% vs. 8.60\%), proving its ability to mitigate the false-positive refusals and verbosity often triggered by restrictive system prompts.


\begin{table}[t]
    \centering
    \scriptsize
    \setlength{\tabcolsep}{4pt}
    \begin{tabular}{l|cc|c|c}
        \toprule
        & \multicolumn{2}{c|}{\textbf{Judged Better}} & & \textbf{Overall (\%)} \\
        \textbf{Model} & \textbf{Safe prompting} & \textbf{\ours{}} & \textbf{Tie} & \textbf{Win+Tie} \\
        \midrule
        Llama-base       & 1,250 & 4,808 & 941   & \textbf{82.14} \\
        Llama-instruct   & 1,087 & 3,702 & 2,210 & \textbf{84.47} \\
        Mistral-base     & 673   & 5,678 & 648   & \textbf{90.38} \\
        Mistral-instruct & 313   & 5,479 & 1,207 & \textbf{95.53} \\
        Qwen-base        & 2,075 & 4,005 & 919   & \textbf{70.35} \\
        Qwen-instruct    & 708   & 5,131 & 1,160 & \textbf{89.88} \\
        Aya-instruct     & 602   & 2,849 & 3,548 & \textbf{91.40} \\
        \midrule
        \textbf{All}     & \textbf{6,708} & \textbf{31,652} & \textbf{10,633} & \textbf{86.31} \\
        \bottomrule
    \end{tabular}
    \caption{\footnotesize Comparative LLM-as-a-Judge evaluation results using Gemini 3.6 Flash. \textbf{Safe prompting} = baseline is better, \textbf{\ours{}} = \ours{} is better. \textbf{Win+Tie} represents the total percentage of prompts where \ours{} was preferred or equally good.}
    \label{tab:comparative_evaluation}
\end{table}



\section{Unsafe rate results}\label{app:unsafe-res}

To complement our comparative evaluations, we measure the absolute reduction in harmful generations using Llama Guard 3 as an automated safety classifier. This unsafe rate metric quantifies the percentage of outputs strictly flagged as 'Unsafe', directly reflecting the intervention's ability to suppress unsafe content across languages.

\paragraph{Takeaways.} 
Across all evaluated models and languages, \ours{} consistently and substantially reduces the rate of unsafe generations, as judged by Llama Guard 3. Unaligned base models exhibit the most improvements, with absolute unsafe rates plummeting by over 50 percentage points in several cases (\textit{e.g.,} Mistral-base and Llama-base). Crucially, the intervention also yields consistent safety gains for instruction-tuned models, such as a 40.12\% absolute reduction for Mistral-instruct in English and up to 14.60\% for Llama-instruct in Japanese, which demonstrates that it effectively patches residual vulnerabilities left by standard safety fine-tuning. The pervasive reductions across all seven target languages confirm that intervening on the shared cross-lingual set of neurons successfully mitigates harm globally, extending robust safety alignment well beyond English.

\begin{table*}[t]
    \centering
    \footnotesize
    \setlength{\tabcolsep}{5pt}
    \begin{tabular}{l|c|c|c|c|c|c|c}
        \toprule
        & \textbf{English}
        & \textbf{Spanish}
        & \textbf{German}
        & \textbf{French}
        & \textbf{Italian}
        & \textbf{Chinese}
        & \textbf{Japanese} \\
        \textbf{Model} & $\Delta$ Unsafe & $\Delta$ Unsafe & $\Delta$ Unsafe & $\Delta$ Unsafe & $\Delta$ Unsafe & $\Delta$ Unsafe & $\Delta$ Unsafe \\
        \midrule
        Llama-base       & -50.70 & -39.25 & -35.14 & -38.02 & -36.28 & -28.70 & -30.86 \\
        Llama-instruct   & -1.05  & -1.66  & -1.92  & -2.80  & -1.75  & -2.27  & -14.60 \\
        Mistral-base     & -56.12 & -53.85 & -52.36 & -48.08 & -52.53 & -38.50 & -36.89 \\
        Mistral-instruct & -40.12 & -32.43 & -26.49 & -27.45 & -26.84 & -26.42 & -28.06 \\
        Qwen-base        & -32.26 & -22.55 & -18.79 & -23.25 & -24.48 & -21.00 & -26.40 \\
        Qwen-instruct    & -4.02  & -2.62  & -3.76  & -5.77  & -2.53  & -2.54  & -5.33  \\
        Aya-instruct     & -3.85  & -4.02  & -3.93  & -6.12  & -6.29  & -4.55  & -10.49 \\
        \bottomrule
    \end{tabular}
    \caption{Unsafe rate difference summary ($\Delta$ Unsafe \%) across seven multilingual model backbones comparing Base vs. \ours{}.}
    \label{tab:unsafe_rate_difference}
\end{table*}



\section{Discussion}
\label{sec:discussion}

\textbf{Effect of \ours{} on the aggression signal}

\noindent The safety scores in Section~\ref{sec:results} measure the outcome of the intervention through an external judge. Here we examine its effect on the model's internal state, asking whether \ours{} actually reduces the aggression signal $\alpha_{S_k}$ (Eq.~\ref{eq:signal}) that it monitors. For each input, we compute $\alpha$ twice: once on the unmodified forward pass (\textit{first pass}) and once with \ours{}'s suppression and steering applied (\textit{second pass}).

\begin{figure*}[t]
    \centering
    \begin{subfigure}{0.24\textwidth}
        \includegraphics[width=\linewidth]{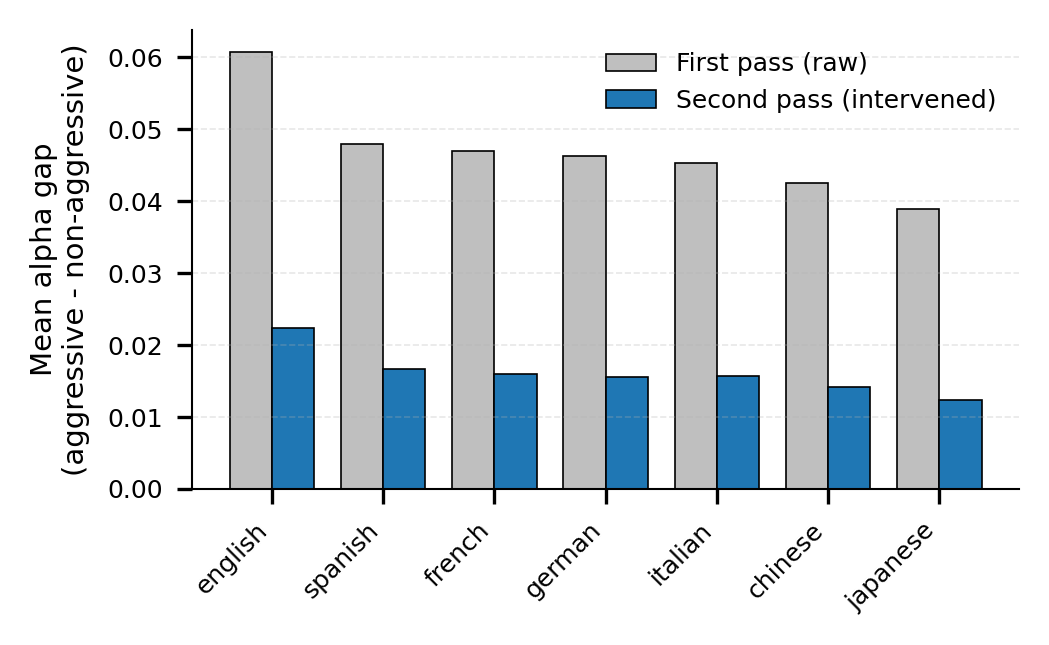}
        \caption{Llama-instruct}
    \end{subfigure}\hfill
    \begin{subfigure}{0.24\textwidth}
        \includegraphics[width=\linewidth]{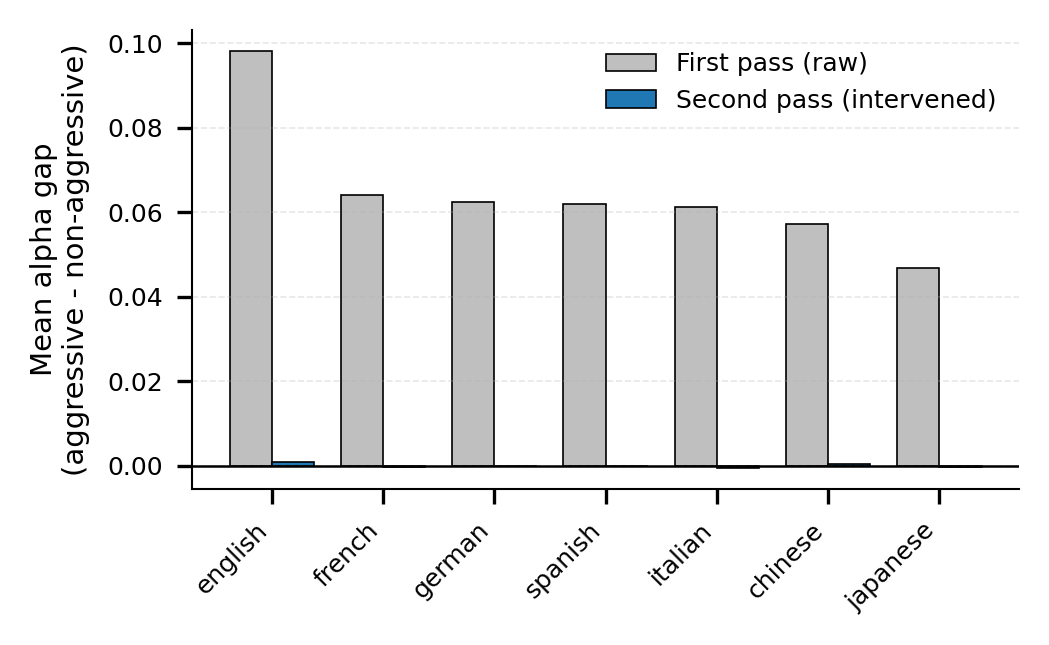}
        \caption{Mistral-instruct}
    \end{subfigure}\hfill
    \begin{subfigure}{0.24\textwidth}
        \includegraphics[width=\linewidth]{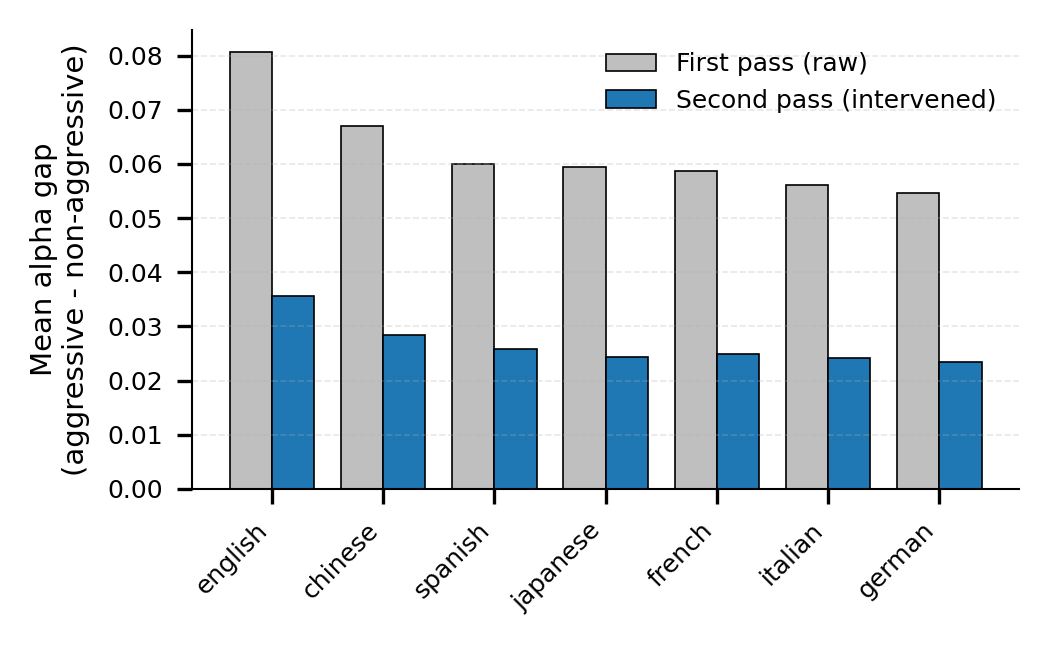}
        \caption{Qwen-instruct}
    \end{subfigure}\hfill
    \begin{subfigure}{0.24\textwidth}
        \includegraphics[width=\linewidth]{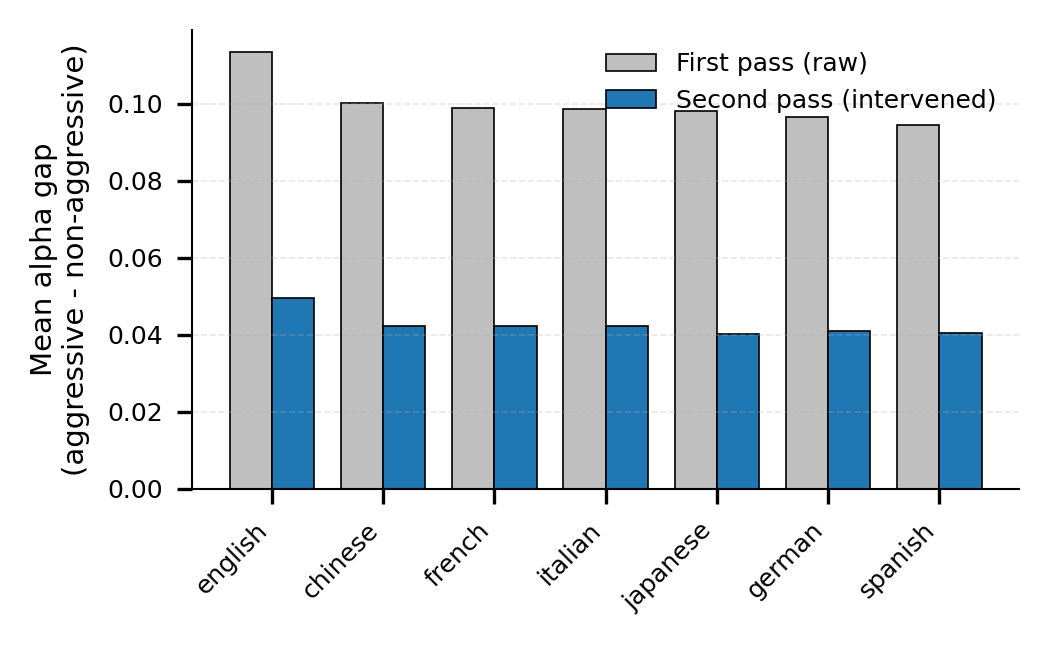}
        \caption{Aya-instruct}
    \end{subfigure}
    \vspace{0.5em}
    \begin{subfigure}{0.24\textwidth}
        \includegraphics[width=\linewidth]{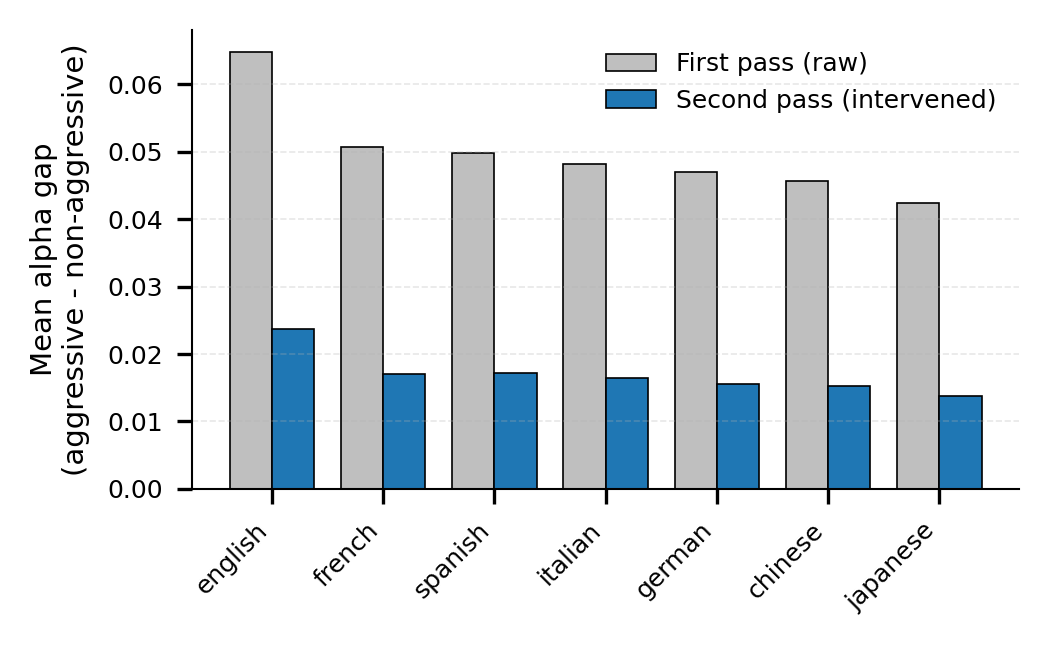}
        \caption{Llama-base}
    \end{subfigure}\hfill
    \begin{subfigure}{0.24\textwidth}
        \includegraphics[width=\linewidth]{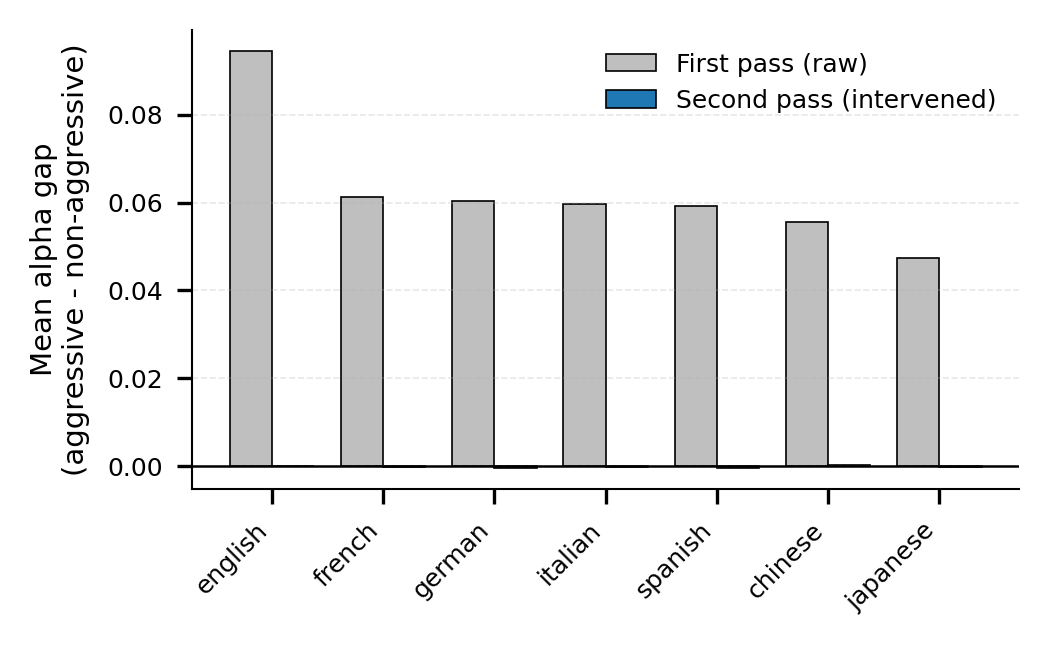}
        \caption{Mistral-base}
    \end{subfigure}\hfill
    \begin{subfigure}{0.24\textwidth}
        \includegraphics[width=\linewidth]{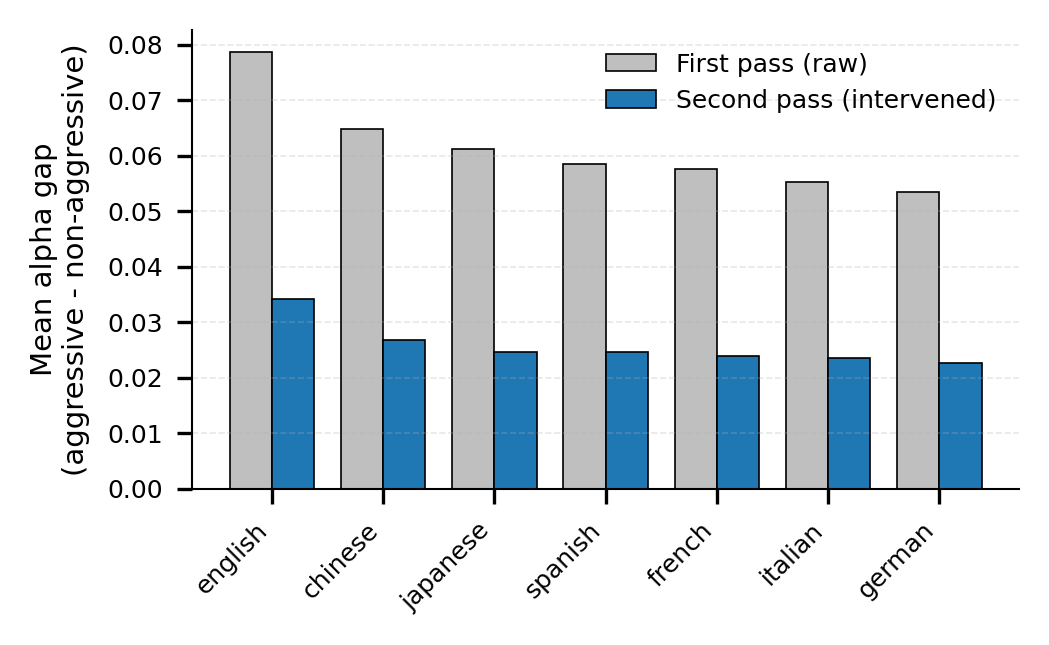}
        \caption{Qwen-base}
    \end{subfigure}
    \caption{Mean $\alpha$ gap (aggressive $-$ non-aggressive) before (grey) and after (blue) \ours{} intervention, across seven languages.}
    \label{fig:alpha_gap}
\end{figure*}

\paragraph{Separation between aggressive and non-aggressive inputs.}
Figure~\ref{fig:alpha_gap} reports the mean $\alpha$ gap, i.e., the difference in mean $\alpha$ between aggressive and non-aggressive inputs, for all six models. Before intervention, the gap is positive for every model and language, ranging from roughly 0.04 to 0.10. English consistently shows the largest gap, in line with it being the source language of both corpora, but the non-English gaps remain of comparable magnitude. After intervention, the gap shrinks substantially in every case. For Mistral-base and Mistral-instruct, it collapses to nearly zero in all languages, meaning that the monitored neurons no longer distinguish aggressive from non-aggressive inputs. Notably, the Mistral models, whose gap is eliminated most completely, also achieve the largest safety improvements among the base and instruct models, respectively (see Table~\ref{tab:cost_per_token}).

\begin{figure*}[t]
    \centering
    \begin{subfigure}{0.24\textwidth}
        \includegraphics[width=\linewidth]{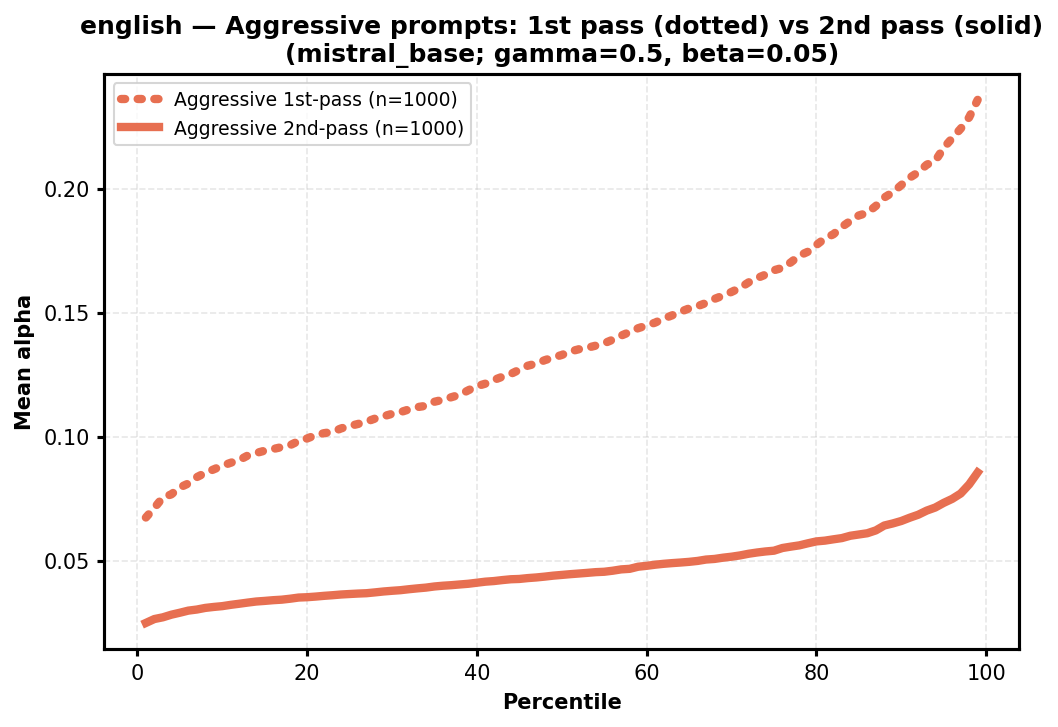}
        \caption{English}
    \end{subfigure}\hfill
    \begin{subfigure}{0.24\textwidth}
        \includegraphics[width=\linewidth]{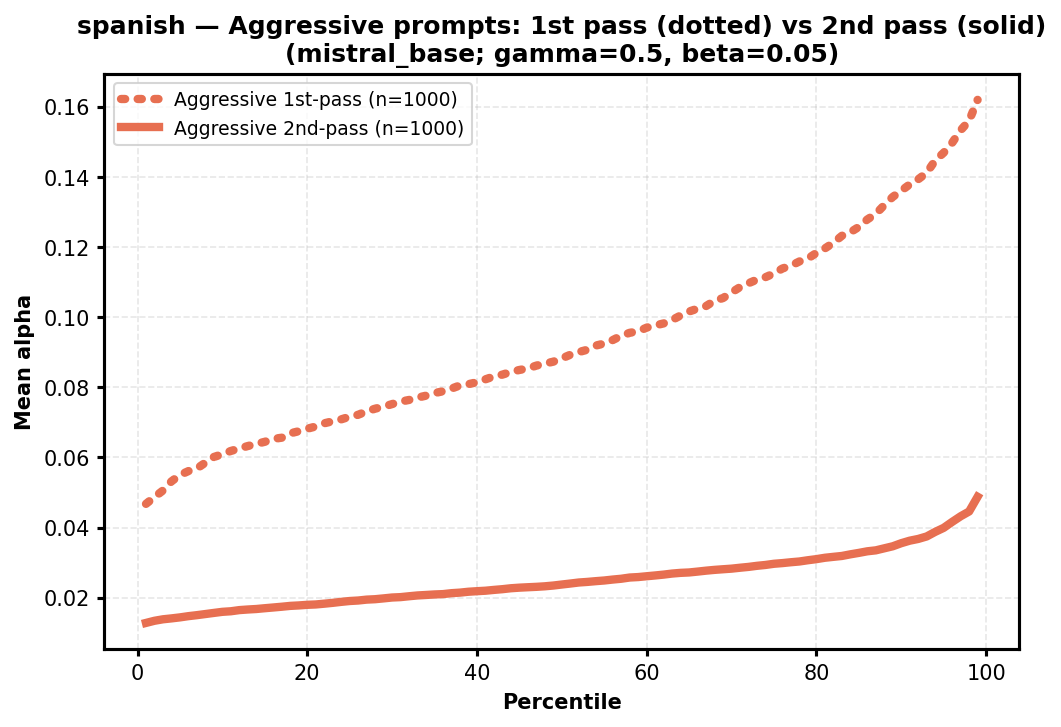}
        \caption{Spanish}
    \end{subfigure}\hfill
    \begin{subfigure}{0.24\textwidth}
        \includegraphics[width=\linewidth]{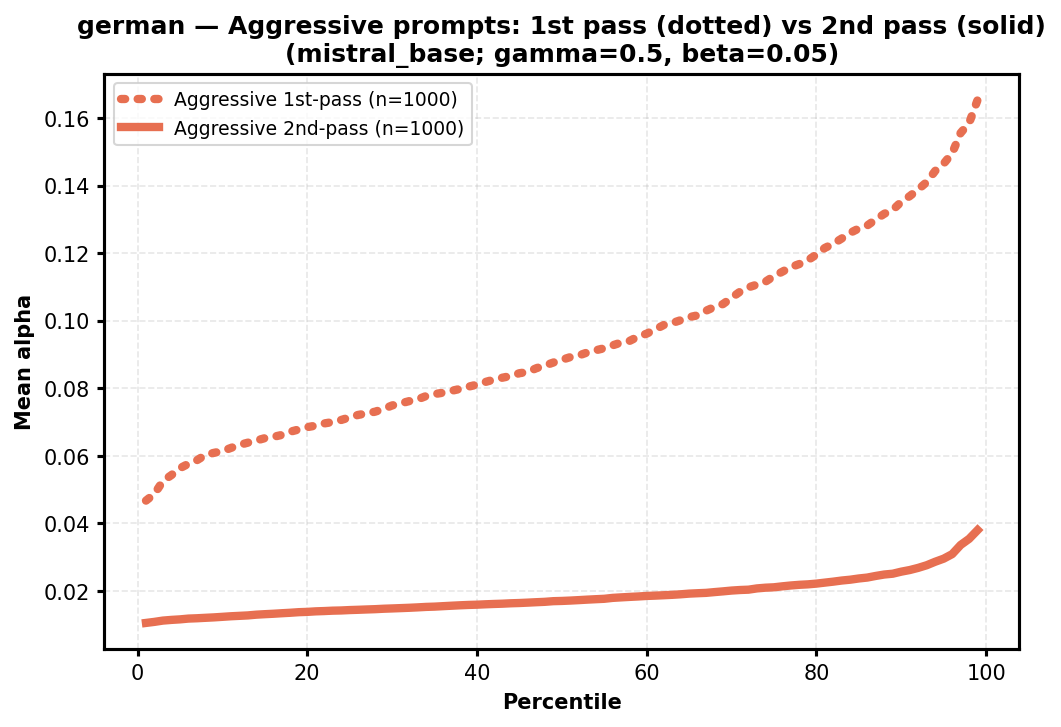}
        \caption{German}
    \end{subfigure}\hfill
    \begin{subfigure}{0.24\textwidth}
        \includegraphics[width=\linewidth]{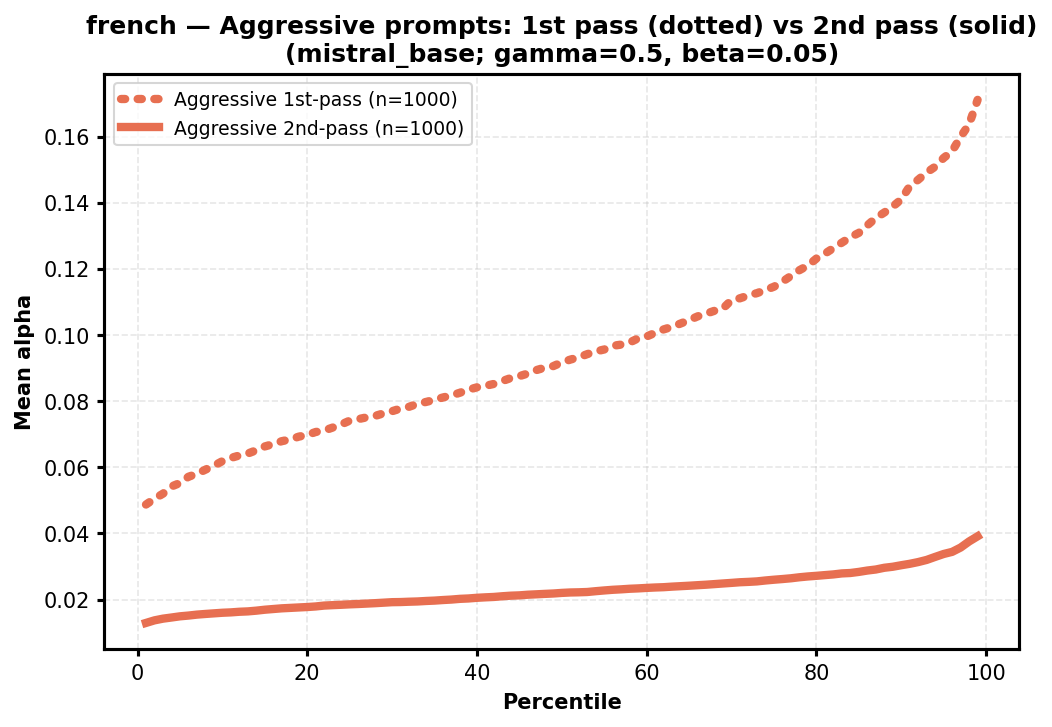}
        \caption{French}
    \end{subfigure}
    \vspace{0.5em}
    \begin{subfigure}{0.24\textwidth}
        \includegraphics[width=\linewidth]{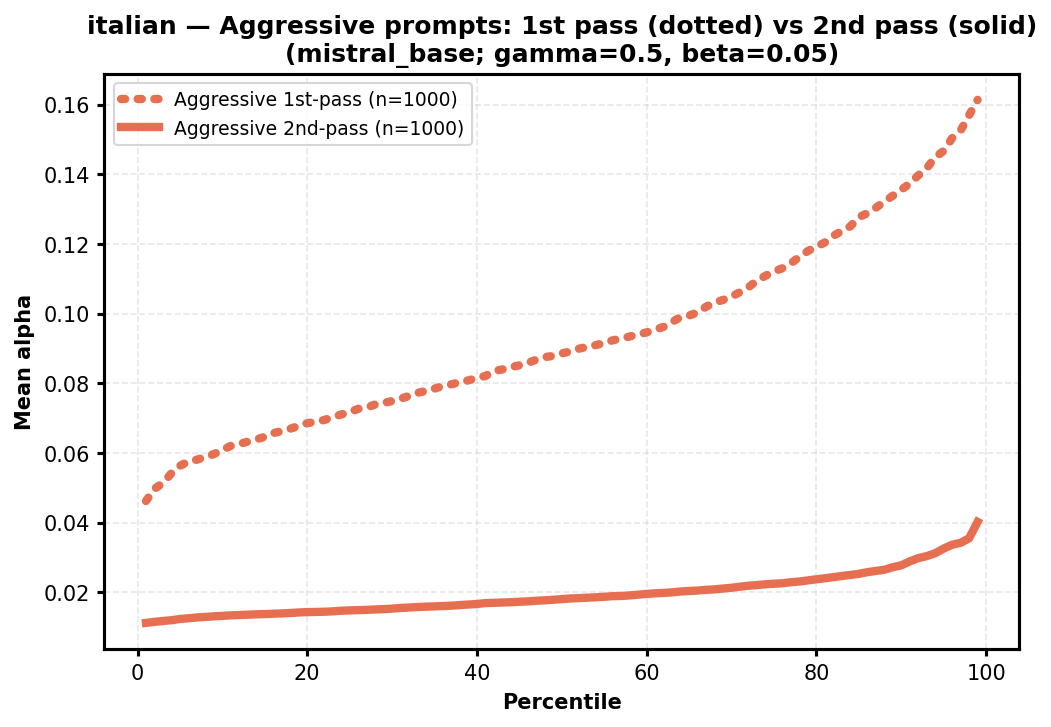}
        \caption{Italian}
    \end{subfigure}\hspace{0.5em}
    \begin{subfigure}{0.24\textwidth}
        \includegraphics[width=\linewidth]{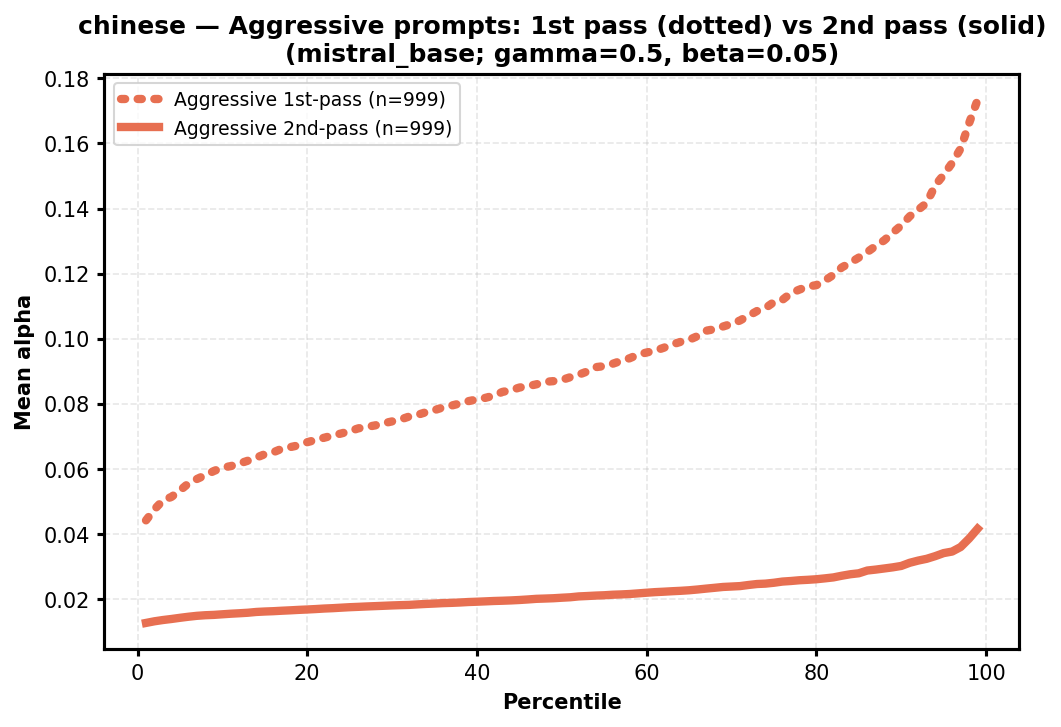}
        \caption{Chinese}
    \end{subfigure}\hspace{0.5em}
    \begin{subfigure}{0.24\textwidth}
        \includegraphics[width=\linewidth]{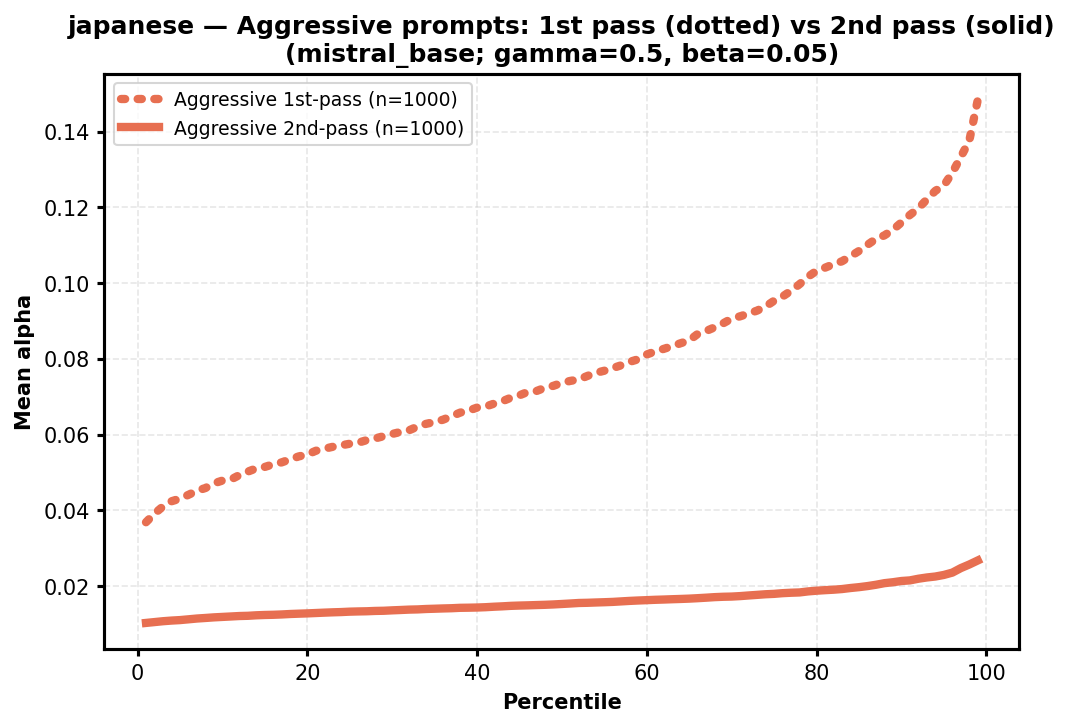}
        \caption{Japanese}
    \end{subfigure}
    \caption{Percentile curves of the aggression signal $\alpha$ for Mistral-base ($\gamma = 0.5$, $\beta = 0.05$). Dotted: first pass (raw); solid: second pass (intervened).}
    \label{fig:alpha_percentile}
\end{figure*}

\paragraph{Distribution of the aggression signal.}
To test whether this reduction is driven by a few inputs or reflects a shift across the entire distribution, Figure~\ref{fig:alpha_percentile} plots the percentile curves of $\alpha$ on aggressive inputs for Mistral-base ($\gamma = 0.5$, $\beta = 0.05$). The intervened curve lies below the raw curve at every percentile. At the median, $\alpha$ drops by roughly two-thirds (from about 0.135 to 0.045 in English), and at the 99th percentile it falls by about 63\% (from 0.235 to 0.086 in English). It is evident that \ours{} shifts the whole distribution downward. The curves also have nearly identical shapes and differ mainly in scale, which again points to a shared mechanism rather than language-specific behaviour.

\section{Computing infrastructure}
\label{app:compute}

All of our experiments were conducted on an NVIDIA standalone machine with $1 \times$ NVIDIA L40 GPU (45 GB VRAM) and an Intel Xeon Silver 4509Y 16-core CPU. The machine has 125 GB of system RAM and runs Ubuntu 22.04.5 LTS with Linux kernel version 5.15.0-170-generic; the models were executed using PyTorch 2.5.1 and the HuggingFace Transformers library version 4.44.2. Tokenization was handled via the tokenizers library (v0.22.1). All models, including
\texttt{meta-llama/Llama-3.1-8B-Instruct}, \texttt{meta-llama/Llama-3.1-8B}, \texttt{mistralai/Mistral-7B-Instruct-v0.3}, \texttt{mistralai/Mistral-7B-v0.3}, \texttt{Qwen/Qwen2.5-7B}, \texttt{Qwen/Qwen2.5-7B-Instruct} and \texttt{CohereLabs/aya-expanse-8b}, were run with a fixed seed 42 to ensure reproducibility and consistency across trials.